%% file: arxiv.tex
\documentclass{article}
\usepackage[T1]{fontenc}
\usepackage[margin=2.8cm]{geometry}
\usepackage{graphicx}
\usepackage{picinpar}
\usepackage{subcaption}
\usepackage{xcolor}
\usepackage{colortbl}
\usepackage{booktabs}
\usepackage{multirow}
\usepackage{soul}
\usepackage{enumerate}
\usepackage{tabularx}

\usepackage{amsmath}
\usepackage{amssymb}
\usepackage{mathtools}
\usepackage{algpseudocode}
\usepackage[ruled,vlined]{algorithm2e}

\usepackage{hyperref}
\usepackage{cleveref}
\usepackage{kpfonts}
\usepackage{natbib}
\usepackage{tcolorbox}
\usepackage{fontawesome5}
\usepackage{xspace}
\usepackage{microtype}
\usepackage{url}
\usepackage{placeins}
\usepackage{authblk}
\usepackage{fancyhdr}
\fancypagestyle{firstpage}{%
  \fancyhead[L]{\raisebox{0.05\height}{\hypersetup{hidelinks}\href{https://www.olaresearch.org/}{\includegraphics[height=1.2em]{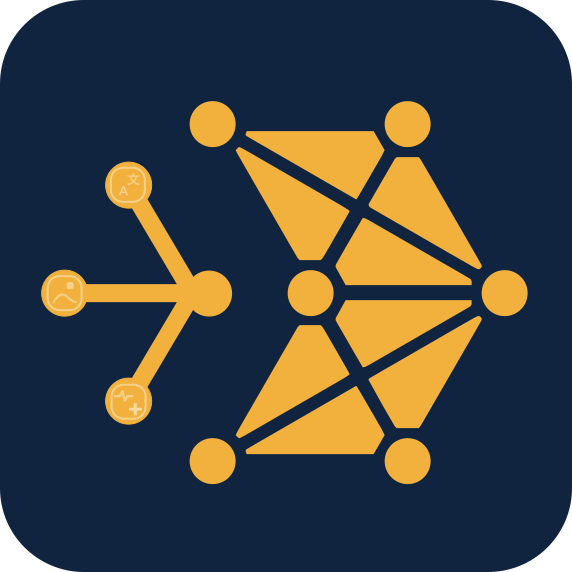}}}~Preprint}%
  \fancyhead[R]{\raisebox{0.05\height}{\hypersetup{hidelinks}\href{https://www.ellisinstitute.fi}{\usebox{\ellislogo}}}}%
}
\newsavebox{\ellislogo}
\sbox{\ellislogo}{\includegraphics[height=1em]{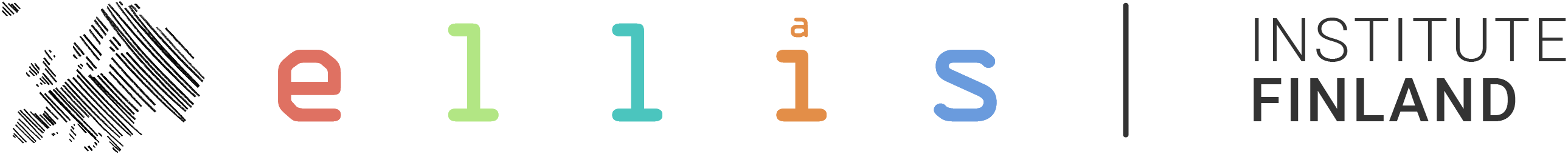}}
\definecolor{colornavy}{HTML}{1B365D}
\definecolor{coloramber}{HTML}{E0913A}
\definecolor{colormint}{HTML}{004A43}
\definecolor{colorblue}{HTML}{1D70B8} \definecolor{algred}{HTML}{C62828} \definecolor{algblue}{HTML}{1D70B8}
\definecolor{colorforest}{HTML}{194D33}
\definecolor{colorslate}{HTML}{475569}
\hypersetup{
  colorlinks=false,
  pdfborder={0 0 1},
  linkbordercolor=colornavy,
  citebordercolor=colorforest,
  urlbordercolor=colorblue,
  pdftitle={MemoryAthena: Adaptive Routing over Latent and Generated Memories},
  pdfauthor={Mingyuan Li, Guangsheng Yu, Juyuan Zhang, Xu Wang, Zhibo Man, Haonan Zhang, Shaoxiong Ji}
}

\setcitestyle{round, elide, numbers}

\renewenvironment{abstract}{%
  \begin{tcolorbox}[
    colback=colornavy!4,
    colframe=colornavy,
    leftrule=4pt,
    rightrule=0.5pt,
    toprule=0.5pt,
    bottomrule=0.5pt,
    arc=3pt,
    boxsep=5pt,
    left=12pt,
    right=12pt,
    top=10pt,
    bottom=10pt,
    title=\textbf{Abstract},
    coltitle=colormint,
    attach title to upper,
    after title={\par\medskip}
  ]%
}{%
  \medskip{\color{colornavy}\hrule height 0.5pt}\medskip
  \small\noindent
  {\hypersetup{hidelinks}%
  \begin{tabular}{@{}l@{\hspace{0.5em}}l@{}}
    \makebox[1.15em][c]{\textcolor{colormint}{\faGlobe}} & \textbf{Project page:} \href{https://olaresearch.org/MemoryATHENA}{\textit{https://olaresearch.org/MemoryATHENA}} \\
    \makebox[1.15em][c]{\textcolor{colormint}{\faGithub}} & \textbf{GitHub code:} \href{https://github.com/OLAResearch/ATHENA}{\textit{https://github.com/OLAResearch/MemoryATHENA}} \\
    \makebox[1.15em][c]{\raisebox{-0.12\height}{\textcolor{colormint}{\faDatabase}}} & \textbf{Hugging Face models:} \href{https://huggingface.co/collections/OLAResearchX/memoryathena}{\textit{https://huggingface.co/collections/OLAResearchX/memoryathena}} 
  \end{tabular}%
  }
  \end{tcolorbox}%
}

\usepackage[]{lineno}

\newcommand{\safeincludegraphics}[2][]{%
  \IfFileExists{#2}{%
    \includegraphics[#1]{#2}%
  }{%
    \fbox{\parbox{0.92\linewidth}{\centering\small
      \textit{Figure omitted: #2 is not included in this source snapshot.}}}%
  }%
}

\definecolor{tablegroup}{HTML}{F3F3F3}
\definecolor{tablefocus}{HTML}{DDF3F8}
\definecolor{gain}{HTML}{007A3D}
\definecolor{loss}{HTML}{C62828}
\newcommand{\groupbar}[2]{\rowcolor{tablegroup}\multicolumn{#1}{c}{\itshape #2}\\}
\newcommand{\avgchange}[2]{#1\,{\scriptsize\ifdim #2pt<0pt\textcolor{loss}{(#2\%)}\else\textcolor{gain}{(#2\%)}\fi}}
\newcommand{\scorechange}[2]{#1\,{\scriptsize\ifdim #2pt<0pt\textcolor{loss}{#2}\else\textcolor{gain}{#2}\fi}}
\definecolor{harmonyblue}{RGB}{222,244,248}
\newcommand{\takeaway}[1]{%
  \par\smallskip\noindent\begingroup
  \setlength{\fboxsep}{6pt}%
  \colorbox{harmonyblue!55}{\parbox{\dimexpr\linewidth-2\fboxsep\relax}{\textbf{Takeaway.} #1}}%
  \endgroup\par\smallskip}

\newcommand{\E}{\mathrm{E}}
\newcommand{\GE}{\mathrm{GE}}
\newcommand{\GH}{\mathrm{GH}}
\newcommand{\clip}{\operatorname{clip}}

\title{MemoryAthena: Adaptive Routing over Latent and Generated Memories}
\author[1,2]{Mingyuan Li} \author[3]{Guangsheng Yu} \author[4]{Juyuan Zhang} \author[3]{Xu Wang} \author[1,2]{Zhibo Man} \author[5]{Haonan Zhang} \author[1,2]{Shaoxiong Ji} \affil[1]{ELLIS Institute of Finland} \affil[2]{University of Turku} \affil[3]{University of Technology Sydney} \affil[4]{University of Science and Technology of China} \affil[5]{Shanghai Jiao Tong University}
\date{}
\begin{document}
\maketitle
\thispagestyle{firstpage}
\begin{abstract}
Learned-memory methods store information in an explicit table and consume it through a separate reader, allowing addressing, storage, and reading to be modified independently. We study whether useful memory can also be generated rather than only retrieved. \textsc{MemoryAthena} uses three pathways: direct Engram retrieval (E), generation from retrieved Engram cues (GE), and generation from causal backbone states without consulting the memory table (GH). Generated memory is conditionally useful: it can complement E in one context but interfere with it in another. \textsc{MemoryAthena} therefore treats E as an anchor and learns when a generated representation should intervene. With the backbone, memory, generators, and readers frozen, a lightweight causal routing head is trained from counterfactual future-token likelihood advantages of GE and GH relative to E. At inference time, an admitted candidate modifies the E residual through bounded interpolation, while rejection recovers the direct pathway exactly. On question answering, \textsc{MemoryAthena} raises the five-task average from 37.65 to 39.28 over the direct pathway of the same checkpoint, while the six-task general-NLP average increases from 76.73 to 79.13. The complete memory-side system contains approximately 201M parameters, excluding the frozen backbone. Further analyses show complementary strengths among E, GE, and GH across tasks and inputs. These results support generated memory as a selective correction to direct retrieval and highlight routing when, which, and how strongly to intervene as the central challenge.
\end{abstract}
\section{Introduction}

Retrieval-augmented generation supplies a language model with external text
\citep{lewis2020rag}, nearest-neighbor language models supply examples drawn
from a non-parametric datastore \citep{khandelwal2020knn}, and learned-memory
approaches supply trainable representations read at inference time
\citep{wei2025mlp,cheng2026engram}.
All three pass a stored item to the model unchanged.
Engram makes that structure explicit by keeping memory outside the backbone,
storing information in an addressable table and consuming it through a
lightweight neural interface \citep{cheng2026engram}.
Prior work treats such a memory as a reusable artifact across language-model
backbones and separates a memory system into three roles
\citep{li2026transfer}.
\emph{Addressing} determines where to access, \emph{memory} stores the
representations, and \emph{reading} transforms a retrieved representation into
a form the backbone can consume.
None of the systems above asks whether a memory representation must be
retrieved from a stored table, or whether useful memory can instead be
generated.

Treating memory as reconstruction rather than literal readout makes the
question tractable. An addressable memory supplies an index or a cue, from
which a neural model can reconstruct a richer internal representation
conditioned on its current context.
The stored memory then need not be the final representation injected into the
model, and is instead a substrate from which a latent memory representation is
generated.
Generation also need not start from an external lookup, because the model's
own causal hidden states may already carry enough contextual evidence to
construct a useful latent representation.
Two forms of generation therefore accompany direct memory access, one
conditioned on retrieved memory cues and one conditioned on the model's causal
context.

Building on the addressing, memory, and reader decomposition of prior work
\citep{li2026transfer}, this paper considers three memory pathways.
Direct Engram retrieval (\textbf{E}) follows the conventional interface and
reads an addressable memory entry unchanged.
Generation from retrieved Engram cues (\textbf{GE}) treats those cues as
conditions for a latent memory representation that the backbone consumes in
its place.
Generation from causal backbone states (\textbf{GH}) forms that representation
without reading the external memory table.
The three differ in how far the representation entering the model departs from
what storage holds, ranging from \emph{direct retrieval} through
\emph{retrieval-conditioned generation} to \emph{context-conditioned
generation}.
GE and GH are alternative memory-side representations attached to the same
frozen backbone rather than additional language models.

Generated memories are not uniformly better than direct retrieval.
A generated representation can be highly useful for one input yet unnecessary
or even harmful for another. This heterogeneity is precisely what makes
generated memory a routing problem rather than a replacement problem: if GE or
GH were consistently superior to E, one could simply replace the direct
pathway. Instead, the useful regime is selective intervention, where a strong
direct-memory pathway is retained and generated memory is introduced only when
it is expected to add value.
The resulting decision is asymmetric. Conventional conditional routing chooses
among several equivalent experts \citep{shazeer2017moe}, whereas here E already
provides a strong direct-memory reference. The model must therefore determine
whether a generated memory should intervene, which generated pathway should be
used, and how strongly it should modify the direct representation. Learning
these decisions from general text rather than downstream task labels makes the
problem particularly challenging.

This paper proposes \textsc{MemoryAthena}, an E-anchored framework for
integrating direct and generated memories. Rather than routing symmetrically
among E, GE, and GH, \textsc{MemoryAthena} treats E as an explicit reference.
A lightweight causal head predicts the E-relative advantage and confidence of
each generated candidate, which is admitted only when both exceed the routing
criteria. The resulting memory residual is
\begin{equation}
    r = e + \alpha(g-e),
    \qquad 0 \leq \alpha \leq 1,
\end{equation}
where $\alpha$ controls the intervention strength. If no generated candidate is
admitted, the model recovers the direct E pathway exactly. Generated memory is
thus treated as a candidate correction rather than a replacement for direct
retrieval.

To train the router without downstream labels, we freeze the backbone, memory,
generators, and readers and compare the E, GE, and GH endpoints under teacher
forcing. For a generated source $s$, we define its token-level advantage over E
as
\begin{equation}
    a_{s,t}
    =
    \log p_s(x_{t+1}\mid x_{\leq t})
    -
    \log p_E(x_{t+1}\mid x_{\leq t}),
\end{equation}
and aggregate these differences over multiple future horizons to supervise the
routing head. Future tokens are used only to construct training targets; at
inference time, routing remains causal. Memory selection therefore becomes an
E-relative utility prediction problem.
Our contributions are threefold:
\begin{itemize}
    \item \textbf{From memory retrieval to memory generation.}
    Building on the decomposition of external memory into addressing,
    storage, and reading \citep{li2026transfer}, we introduce a three-path memory
    interface spanning direct retrieval (E), generation from retrieved Engram
    cues (GE), and generation from causal backbone states (GH).
    We use it to investigate whether the representation consumed by a language
    model must be explicitly stored, or can instead be generated from memory
    cues or contextual states.

    \item \textbf{Routing under heterogeneous memory utility.}
    We formulate generated memory as a conditional intervention problem: GE and GH
    can complement direct retrieval, but neither is uniformly preferable.
    \textsc{MemoryAthena} therefore retains E as an explicit reference and learns
    when, which, and how strongly a generated representation should intervene,
    using bounded interpolation with exact fallback to the direct pathway.

    \item \textbf{Counterfactual advantage distillation without downstream
    supervision.}
    We construct routing targets from future-token likelihood differences among
    frozen memory pathways and distill them into a lightweight causal head.
    This separates learning \emph{how to construct} memory representations from
    learning \emph{when to use} them.
\end{itemize}

We evaluate \textsc{MemoryAthena} on question answering and general NLP tasks.
It improves all five QA summary metrics and five of six NLP tasks over the
direct-memory pathway of the same checkpoint.
Analyses of the individual endpoints reveal complementary successful
predictions across E, GE, and GH.

\section{Background and Problem Setup}
\label{sec:background}

\textbf{External and learned memory.}
A language model can reach information its backbone parameters do not hold, by
retrieval or by learned memory. Retrieval-augmented generation conditions the
output on retrieved text \citep{lewis2020rag}, while $k$NN-LMs interpolate
neural predictions with a distribution drawn from a nearest-neighbor datastore
\citep{khandelwal2020knn}. Learned-memory approaches move the memory into trained representations
instead, so that MLP Memory learns a parametric memory module
\citep{wei2025mlp} and Engram introduces an addressable conditional memory
based on causal $N$-gram lookup \citep{cheng2026engram}. Engram is the direct memory substrate
throughout, and the one thing varied here is how the representation consumed by
the backbone is constructed from that stored memory or built beside it.

Structured control of model outputs has also been studied outside language memory. PreResQ-R1 disentangles response stability from preference learning for visual quality assessment \citep{feng2025preresqr1}, while TriMRG uses triple-graph semantic reasoning to reduce hallucinations in medical report generation \citep{feng2026trimrg}. These systems motivate explicit mechanisms that regulate generated representations instead of accepting every generated state unchanged.

\textbf{Reusable memory and target-side reading.}
Cross-model memory transfer separates a memory system into three roles
\citep{li2026transfer}. \emph{Addressing} determines what memory is accessed,
\emph{memory} holds the reusable representations, and \emph{reading} adapts a
retrieved representation to the target backbone. A memory learned with one
language model can then remain frozen and be reused by another backbone through
an adapted target-side reader. The stored representation and the representation the model finally consumes
therefore need not be identical, which is the property this paper builds on.

Let $x_{1:T}$ be a token sequence and $f_\theta$ a frozen autoregressive
backbone. An addressable memory $M_\phi$ retrieves
\begin{equation}
    m_t = M_\phi[\operatorname{canon}(x_{\leq t})],
\end{equation}
where $\operatorname{canon}(\cdot)$ denotes the canonical addressing rule.
A reader then maps the retrieved memory and the current hidden state
$h_t^\ell$ into a residual contribution that is injected as
\begin{equation}
    h_t^\ell \leftarrow h_t^\ell + r_t^\ell .
\end{equation}
\textsc{MemoryAthena} starts from this addressing, memory and reader view.

\textbf{From memory reading to memory generation.}
If the representation consumed by the backbone is already produced through a
learned interface, it need not be obtained by reading the stored memory
directly, and three alternatives follow. The \textbf{E} pathway reads the retrieved Engram
representation directly and produces a residual $e_t^\ell$.
The \textbf{GE} pathway generates a latent memory representation conditioned on
retrieved Engram cues, producing $g_{\mathrm{GE},t}^\ell$.
The \textbf{GH} pathway generates a latent memory representation from causal
backbone states without consulting the external memory table, producing
$g_{\mathrm{GH},t}^\ell$.
All three are residual representations in the same target hidden-state space,
attached to the same frozen backbone.

For $s\in\{\mathrm{E},\mathrm{GE},\mathrm{GH}\}$, we denote by $p_s$ the
endpoint distribution obtained when pathway $s$ is used throughout the
configured memory-injection sites, and these endpoints are the common reference
against which direct and generated memory representations are compared.

\textbf{Conditional routing over memory representations.}
Mixture-of-experts
methods learn input-dependent combinations of expert outputs
\citep{shazeer2017moe,fedus2022switch}, and memory-augmented language models
have used learned selection mechanisms \citep{merity2017pointer}. Our setting
is asymmetric. E is already a usable direct-memory pathway, whereas GE and GH
are candidate modifications to that reference. The decision is therefore
whether a generated representation provides additional utility over E, which
candidate should intervene, and how strongly it should modify the direct
residual.
Section~\ref{sec:method} develops the routing mechanism for this E-relative
decision.

\begin{figure}[ht]
    \centering
    \safeincludegraphics[width=\columnwidth]{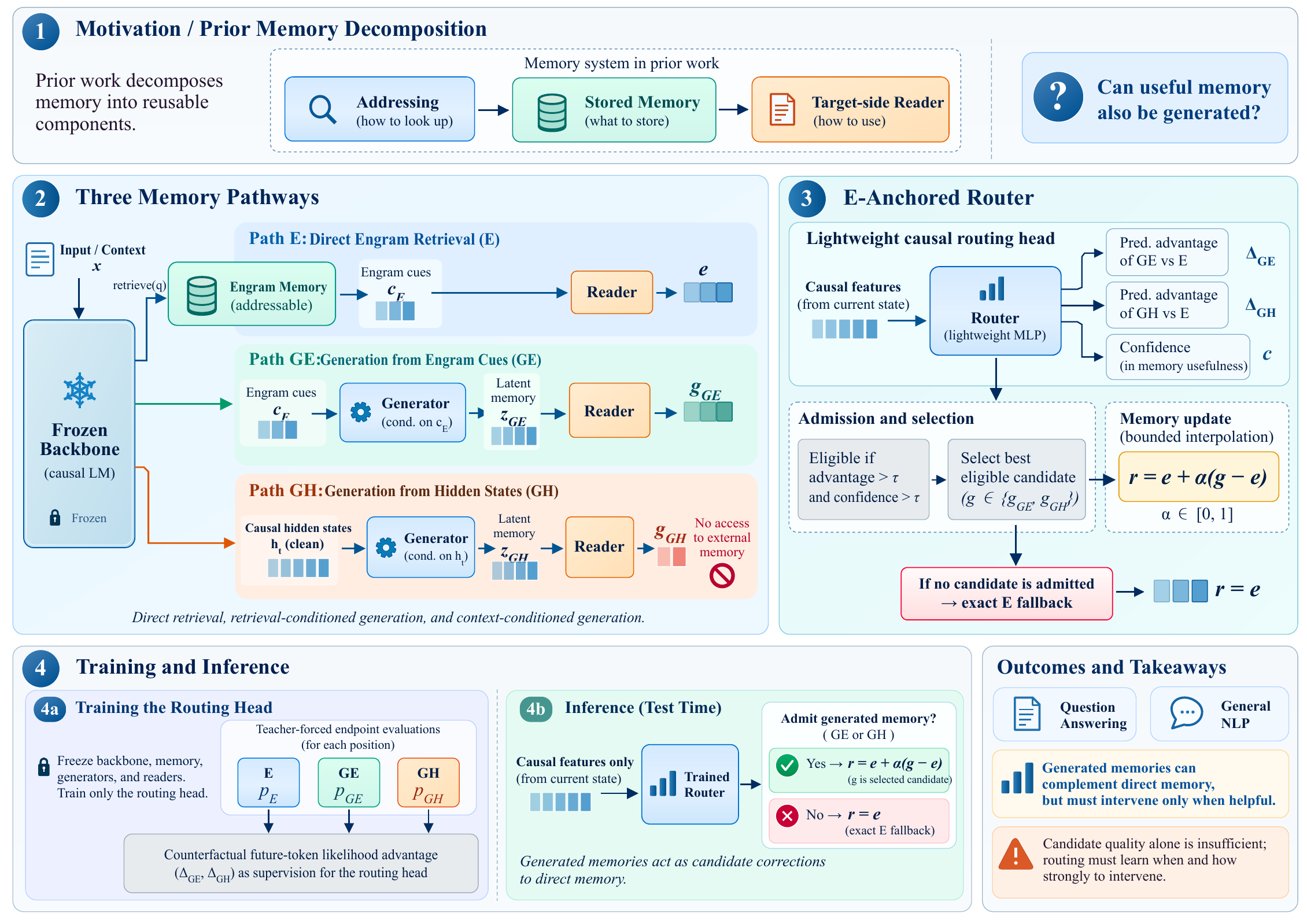}

    \caption{Overview of \textsc{MemoryAthena}. The framework extends direct Engram retrieval (E) with two generated-memory pathways, generation from retrieved Engram cues (GE) and generation from causal backbone states (GH). A lightweight E-anchored router predicts the relative advantage of GE and GH, conditionally admits a generated candidate, and controls its contribution through bounded interpolation with exact fallback to E. The routing head is trained from counterfactual future-token likelihood differences while the memory pathways remain frozen.}
    \label{fig:transfer-overview}
\end{figure}

\section{METHOD}
\label{sec:method}
\textsc{MemoryAthena} separates memory construction from memory selection.
Training proceeds in three stages. First, an addressable memory is learned under
causal language-modeling supervision while the source backbone is frozen.
Second, the memory and target backbone are fixed, and the memory-side interfaces
are adapted to construct the direct pathway E and the generated pathways GE and
GH. Third, all memory pathways are frozen and only a lightweight routing head is
trained to predict the E-relative utility of GE and GH from counterfactual
future-token supervision. The model therefore first learns \emph{how to
construct candidate memory representations} and then learns \emph{when and
how strongly a generated representation should modify the direct memory}.
Detailed objectives, architectures, and optimization settings are provided in
Appendix~\ref{app:implementation}.

\subsection{Stage1: Direct and Generated Memory Pathways}
\label{sec:pathways}

We consider a frozen autoregressive backbone augmented with three
memory-side pathways that differ in how the representation injected into the
backbone is constructed.
Let $h_t^\ell$ denote the backbone hidden state at position $t$ and injection
layer $\ell$, and let $m_t$ denote the representation retrieved from the
addressable Engram memory.
The three pathways produce residual contributions in the same target hidden
space:
\begin{align}
e_t^\ell
&=
R_{\mathrm{E}}^\ell
\left(h_t^\ell, m_t\right),
\\
g_{\mathrm{GE},t}^\ell
&=
R_{\mathrm{GE}}^\ell
\left(h_t^\ell, z_{\mathrm{E},t}^\ell\right),
\\
g_{\mathrm{GH},t}^\ell
&=
R_{\mathrm{GH}}^\ell
\left(h_t^\ell, z_{\mathrm{H},t}^\ell\right).
\end{align}

The \textbf{E} pathway directly reads the retrieved Engram representation.
The \textbf{GE} pathway first generates a latent memory representation
$z_{\mathrm{E},t}^\ell$ conditioned on retrieved Engram cues and then maps it
into the backbone hidden space.
The \textbf{GH} pathway instead generates
$z_{\mathrm{H},t}^\ell$ from causal backbone states obtained in a separate
pass with memory injection disabled.

GE and GH are memory-side representations rather than independent language
models, and all three pathways operate around the same frozen backbone.

\subsection{Stage2: E-Relative Advantage Distillation}
\label{sec:advantage}

Generated memory is not uniformly preferable to direct memory.
We therefore treat the direct E pathway as an explicit reference and learn
whether each generated candidate is expected to improve upon it.
After the memory pathways have been learned, we freeze the backbone, memory,
generators and readers.
For each endpoint
$s \in \{\mathrm{E},\mathrm{GE},\mathrm{GH}\}$,
we obtain a next-token distribution $p_s$ by using that pathway alone under
teacher forcing.
For a generated source
$s \in \{\mathrm{GE},\mathrm{GH}\}$,
we define its token-level advantage relative to E as
\begin{equation}
a_{s,t}
=
\log p_s
\left(
x_{t+1} \mid x_{\leq t}
\right)
-
\log p_{\mathrm{E}}
\left(
x_{t+1} \mid x_{\leq t}
\right).
\label{eq:token_advantage}
\end{equation}
A positive value indicates that the generated pathway assigns greater
likelihood to the observed future token than the direct-memory pathway.
Token-level advantages are aggregated over multiple future positions to
obtain a smoother training target, denoted by $A_{s,t}$.
A lightweight routing head predicts an advantage and a confidence score $c_{s,t}$ for
each generated candidate:
\begin{equation}
\left(
\widehat{A}_{s,t}^{\ell},
c_{s,t}^{\ell}
\right)
=
q^\ell
\left(
\operatorname{stopgrad}
\left(
\phi_{s,t}^{\ell}
\right)
\right),
\label{eq:router_prediction}
\end{equation}
where $\phi_{s,t}^{\ell}$ denotes current-state features available up to position $t$, including the backbone hidden state and statistics of the direct and generated memory residuals. The scalar $\widehat{A}_{s,t}^{\ell}$ predicts how much pathway $s$ is expected to improve over E, while $c_{s,t}^{\ell}$ provides an additional confidence signal for admission. The $\operatorname{stopgrad}$ operator keeps the routing loss from updating the backbone or the memory pathways. Only $q^\ell$ is trained. Future tokens enter only the E-relative advantage targets during training, and at inference time the router relies on current-state features alone. Appendix~\ref{app:implementation} specifies the feature set $\phi$ and the routing objective.

\begin{algorithm}[!htbp]
\caption{\textsc{MemoryAthena}}
\label{alg:memoryathena}
\small

\KwIn{Direct residual $e$; generated residuals $g_{\mathrm{GE}}$ and
$g_{\mathrm{GH}}$; routing head $q$; thresholds $\tau$ and $\rho$;
maximum scale $a_{\max}$; temperature $T_{\alpha}$}

\KwOut{Memory residual $r$}

\BlankLine
\tcp{\textbf{// Predict E-relative advantage}}
$(\widehat{A}_{s},c_s)_{s\in\{\mathrm{GE},\mathrm{GH}\}}
\leftarrow q(\text{current-state features})$;

\BlankLine
\tcp{\textcolor{algred}{\textbf{// Admit only beneficial generated memories}}}

$\mathcal{C} \leftarrow
\{\,s\in\{\mathrm{GE},\mathrm{GH}\} :
\widehat{A}_{s}>\tau
\;\wedge\;
\sigma(c_s)\geq\rho\,\}$;

\If{$\mathcal{C}=\varnothing$}{
\textcolor{algred}{\textbf{return} $e$}
\hfill
\textit{[exact E fallback]};
}

\BlankLine
\tcp{\textcolor{algblue}{\textbf{// Select the best generated candidate}}}

$s^{\star}
\leftarrow
\displaystyle\arg\max_{s\in\mathcal{C}}
\widehat{A}_{s}$;

\BlankLine
\tcp{\textcolor{algblue}{\textbf{// Determine intervention strength}}}

$\alpha
\leftarrow
a_{\max}
\operatorname{clip}
\left(
\frac{\widehat{A}_{s^{\star}}-\tau}{T_{\alpha}},
0,1
\right)
\sigma(c_{s^{\star}})$;

\BlankLine
\tcp{\textcolor{algred}{\textbf{// Apply an E-anchored correction}}}

$r
\leftarrow
e+\alpha\bigl(g_{s^{\star}}-e\bigr)$;

\textbf{return} $r$;

\end{algorithm}

\subsection{Stage3: E-Anchored Memory Routing}
\label{sec:routing}

At inference time, GE and GH are treated as candidate corrections to the
direct E pathway rather than as symmetric experts.

For each generated source
$s \in \{\mathrm{GE},\mathrm{GH}\}$,
we test whether its predicted advantage and confidence satisfy the admission
criteria:
\begin{equation}
\mathcal{C}_t^\ell
=
\left\{
s \in \{\mathrm{GE},\mathrm{GH}\}
\; \middle| \;
\widehat{A}_{s,t}^{\ell} > \tau,
\quad
\sigma\!\left(c_{s,t}^{\ell}\right) \geq \rho
\right\},
\label{eq:eligible_set}
\end{equation}
where $\tau$ and $\rho$ are the advantage and confidence thresholds.
If at least one generated source is eligible, the router selects the candidate
with the largest predicted advantage:
$
s^\star
=
\arg\max_{s \in \mathcal{C}_t^\ell}
\widehat{A}_{s,t}^{\ell}.
$
The selected generated representation modifies the direct residual through
bounded interpolation:
\begin{equation}
r_t^\ell
=
e_t^\ell
+
\alpha_t^\ell
\left(
g_{s^\star,t}^\ell - e_t^\ell
\right),
\qquad
0 \leq \alpha_t^\ell \leq 1.
\label{eq:anchored_interpolation}
\end{equation}

The interpolation strength $\alpha_t^\ell$ increases with the predicted utility and confidence of the selected candidate. Algorithm~\ref{alg:memoryathena} uses maximum scale $a_{\max}=1$ and temperature $T_\alpha=0.15$.
If no generated source is admitted, we set $\alpha_t^\ell = 0$ and then
$r_t^\ell = e_t^\ell$.

The resulting residual is injected into the backbone as
$h_t^\ell
\leftarrow
h_t^\ell + r_t^\ell.$
E therefore holds a privileged role. Generated memory modifies the
direct-memory contribution only when it is predicted to be useful, and
rejection recovers the direct E pathway exactly at the corresponding injection
site.

\section{Experiments}

We evaluate whether generated memory improves a direct reader, whether its pathways provide complementary answers, and whether E-relative admission and the reader interface explain the gains. The primary backbone is Mistral-7B-v0.3~\citep{jiang2023mistral}, with memory injected at layers 2 and 10 through a four-branch reader following \citet{li2026transfer}. Llama-2-7B~\citep{touvron2023llama} supplies the imported source memory and is the target backbone in the cross-backbone transfer row. We compare frozen inference rules within a shared checkpoint on five QA benchmarks and six classification tasks. Dataset definitions, scoring, sample counts, and training budgets appear in Appendices~\ref{app:setup} and~\ref{app:training}.
\begin{table}[t]
\centering
\caption{
QA comparison, routing-policy analysis, and cross-backbone transfer on Natural
Questions (NQ), WebQuestions (WebQA), TriviaQA, TruthfulQA and HotpotQA.
Average is the unweighted mean of four F1 scores and one TruthfulQA
multiple-choice (MC) summary.
}
\label{tab:qa_combined}
\small
\setlength{\tabcolsep}{3.5pt}

\resizebox{\linewidth}{!}{%
\begin{tabular}{lrrrrrr}
\toprule
Method / Inference rule
& NQ $\uparrow$
& WebQA $\uparrow$
& TriviaQA $\uparrow$
& TruthfulQA $\uparrow$
& HotpotQA $\uparrow$
& Average $\uparrow$\\
\midrule

\groupbar{7}{Reported baselines}

Base (Vanilla Mistral)
& 20.60
& 29.30
& 57.70
& 32.10
& 21.00
& 32.14\\

RAG
& \scorechange{22.60}{+2.00}
& \scorechange{24.90}{-4.40}
& \scorechange{54.20}{-3.50}
& \scorechange{35.50}{+3.40}
& \scorechange{29.80}{+8.80}
& \avgchange{33.40}{+3.9}\\

kNN-LM
& \scorechange{21.10}{+0.50}
& \scorechange{30.50}{+1.20}
& \scorechange{57.80}{+0.10}
& \scorechange{32.30}{+0.20}
& \scorechange{21.20}{+0.20}
& \avgchange{32.58}{+1.4}\\

CPT
& \scorechange{12.20}{-8.40}
& \scorechange{34.10}{+4.80}
& \scorechange{61.20}{+3.50}
& \scorechange{29.20}{-2.90}
& \scorechange{16.00}{-5.00}
& \avgchange{30.54}{-5.0}\\

LoRA
& \scorechange{18.20}{-2.40}
& \scorechange{34.50}{+5.20}
& \scorechange{61.60}{+3.90}
& \scorechange{30.90}{-1.20}
& \scorechange{16.20}{-4.80}
& \avgchange{32.28}{+0.4}\\

MLP Memory
& \scorechange{25.20}{+4.60}
& \scorechange{37.50}{+8.20}
& \scorechange{61.00}{+3.30}
& \scorechange{32.50}{+0.40}
& \scorechange{24.10}{+3.10}
& \avgchange{36.06}{+12.2}\\

\midrule
\groupbar{7}{Shared-checkpoint memory pathways}

E only
& \scorechange{28.28}{+7.68}
& \scorechange{33.35}{+4.05}
& \scorechange{69.34}{+11.64}
& \scorechange{31.25}{-0.85}
& \scorechange{26.04}{+5.04}
& \avgchange{37.65}{+17.1}\\

GE only
& \scorechange{28.75}{+8.15}
& \scorechange{35.24}{+5.94}
& \scorechange{58.62}{+0.92}
& \scorechange{31.14}{-0.96}
& \scorechange{23.18}{+2.18}
& \avgchange{35.39}{+10.1}\\

GH only
& \scorechange{24.55}{+3.95}
& \scorechange{30.80}{+1.50}
& \scorechange{56.70}{-1.00}
& \scorechange{31.04}{-1.06}
& \scorechange{22.95}{+1.95}
& \avgchange{33.21}{+3.3}\\

\midrule
\groupbar{7}{Alternative routing and fusion rules}

Ordinary hard routing
& \scorechange{26.69}{+6.09}
& \scorechange{35.13}{+5.83}
& \scorechange{61.80}{+4.10}
& \scorechange{31.18}{-0.92}
& \scorechange{24.36}{+3.36}
& \avgchange{35.83}{+11.5}\\

Ordinary soft fusion
& \scorechange{19.66}{-0.94}
& \scorechange{26.05}{-3.25}
& \scorechange{46.36}{-11.34}
& \scorechange{31.13}{-0.97}
& \scorechange{19.91}{-1.09}
& \avgchange{28.62}{-10.9}\\

Subset hard routing
& \scorechange{28.42}{+7.82}
& \scorechange{33.30}{+4.00}
& \scorechange{69.38}{+11.68}
& \scorechange{31.55}{-0.55}
& \scorechange{26.06}{+5.06}
& \avgchange{37.74}{+17.4}\\

Subset soft fusion
& \scorechange{23.17}{+2.57}
& \scorechange{28.71}{-0.59}
& \scorechange{44.37}{-13.33}
& \scorechange{31.89}{-0.21}
& \scorechange{19.27}{-1.73}
& \avgchange{29.48}{-8.3}\\

\midrule
\groupbar{7}{Cross-backbone transfer and E-anchored routing}

Mistral Zero-shot $\rightarrow$ Llama
& \scorechange{29.98}{+9.38}
& \scorechange{36.40}{+7.10}
& \scorechange{67.39}{+9.69}
& \scorechange{30.41}{-1.69}
& \scorechange{25.17}{+4.17}
& \avgchange{37.87}{+17.8}\\

\rowcolor{tablefocus}
\textbf{\textsc{MemoryAthena}}
& \scorechange{\textbf{33.02}}{+12.42}
& \scorechange{\textbf{34.60}}{+5.30}
& \scorechange{\textbf{70.68}}{+12.98}
& \scorechange{\textbf{31.74}}{-0.36}
& \scorechange{\textbf{26.34}}{+5.34}
& \avgchange{\textbf{39.28}}{+22.2}\\

\bottomrule
\end{tabular}%
}
\par\smallskip
\begin{minipage}{\linewidth}
\end{minipage}

\end{table}
\begin{table}[t]
\centering
\caption{
Six-task NLP accuracy (\%) on SST2, MR, CR, RT, AG News (AGN) and Yahoo
Answers. Small signed values are percentage-point differences
from the Full-choice dCPMI Base.
}
\label{tab:nlp}
\small
\setlength{\tabcolsep}{3pt}

\resizebox{\linewidth}{!}{%
\begin{tabular}{lrrrrrrr}
\toprule
Method
& SST2 $\uparrow$
& MR $\uparrow$
& CR $\uparrow$
& RT $\uparrow$
& AGN $\uparrow$
& Yahoo $\uparrow$
& Average $\uparrow$\\
\midrule

\groupbar{8}{Full-choice dCPMI baseline}
Mistral-7B-v0.3
& 81.08
& 75.60
& 74.00
& 74.67
& 73.24
& 55.03
& 72.27\\

\groupbar{8}{Non-parametric methods (reported)}
RAG
& \scorechange{87.20}{+6.12}
& \scorechange{83.70}{+8.10}
& \scorechange{71.55}{-2.45}
& \scorechange{82.36}{+7.69}
& \scorechange{75.64}{+2.40}
& \scorechange{58.43}{+3.40}
& \avgchange{76.48}{+5.8}\\

kNN-LM
& \scorechange{82.15}{+1.07}
& \scorechange{76.85}{+1.25}
& \scorechange{61.70}{-12.30}
& \scorechange{74.95}{+0.28}
& \scorechange{76.13}{+2.89}
& \scorechange{56.26}{+1.23}
& \avgchange{71.34}{-1.3}\\

\groupbar{8}{Parametric methods (reported)}
CPT
& \scorechange{87.09}{+6.01}
& \scorechange{82.85}{+7.25}
& \scorechange{82.60}{+8.60}
& \scorechange{77.48}{+2.81}
& \scorechange{83.10}{+9.86}
& \scorechange{51.56}{-3.47}
& \avgchange{77.45}{+7.2}\\

LoRA
& \scorechange{86.54}{+5.46}
& \scorechange{83.20}{+7.60}
& \scorechange{75.10}{+1.10}
& \scorechange{79.83}{+5.16}
& \scorechange{65.46}{-7.78}
& \scorechange{57.30}{+2.27}
& \avgchange{74.57}{+3.2}\\

MLP Memory
& \scorechange{83.19}{+2.11}
& \scorechange{79.90}{+4.30}
& \scorechange{75.95}{+1.95}
& \scorechange{75.42}{+0.75}
& \scorechange{80.28}{+7.04}
& \scorechange{57.33}{+2.30}
& \avgchange{75.35}{+4.3}\\

\groupbar{8}{This study}
E only
& \scorechange{84.17}{+3.09}
& \scorechange{81.00}{+5.40}
& \scorechange{82.40}{+8.40}
& \scorechange{82.36}{+7.69}
& \scorechange{72.93}{-0.31}
& \scorechange{57.51}{+2.48}
& \avgchange{76.73}{+6.2}\\

\rowcolor{tablefocus}
\textbf{\textsc{MemoryAthena}}
& \scorechange{\textbf{88.07}}{+6.99}
& \scorechange{\textbf{84.70}}{+9.10}
& \scorechange{\textbf{84.10}}{+10.10}
& \scorechange{\textbf{83.86}}{+9.19}
& \scorechange{\textbf{76.64}}{+3.40}
& \scorechange{\textbf{57.43}}{+2.40}
& \avgchange{\textbf{79.13}}{+9.5}\\

\bottomrule
\end{tabular}%
}

\par\smallskip
\begin{minipage}{\linewidth}
\end{minipage}

\end{table}

\subsection{RQ1: When does generated memory improve a strong direct-memory pathway?}
\label{sec:rq1}

Table~\ref{tab:qa_combined} reports the QA results
together with literature baselines, individual pathways, alternative routing
rules and cross-backbone transfer, and Table~\ref{tab:nlp} reports the six
general NLP tasks.

\textbf{Compared routing strategies.}
\emph{E only}, \emph{GE only}, and \emph{GH only} force one memory pathway
throughout inference.
\emph{Ordinary hard routing} treats the three pathways as symmetric candidates
and selects a single pathway, while \emph{ordinary soft fusion} combines their
representations using learned routing weights.
The \emph{subset hard} and \emph{subset soft} variants restrict routing to a
learned subset of sources. The hard variant makes a discrete routing decision
within that subset, whereas the soft variant fuses the subset with learned
weights.
\textsc{MemoryAthena} instead treats E as the reference pathway throughout.
GE or GH modifies E only when the predicted E-relative advantage and
confidence satisfy the admission criteria, and the selected representation is
combined with E through bounded interpolation. Appendix~\ref{app:implementation}
defines the subset-routing controls.

\textbf{QA performance.}
The individual pathways show that generated memory is useful but not uniformly
better than direct retrieval. GE improves WebQA from 33.35 to 35.24 but is
weaker than E on TriviaQA and HotpotQA, and GH is weaker than E on all five QA
summary metrics. This makes unconditional replacement of E undesirable.

The alternative routing rules lead to the same conclusion. Ordinary hard
routing reaches an average of 35.83, while ordinary soft fusion falls to 28.62.
The stronger subset-hard control reaches 37.74, but remains below
\textsc{MemoryAthena} at 39.28. Relative to the same-checkpoint E pathway,
\textsc{MemoryAthena} improves all five QA metrics, by 4.74 points on NQ,
1.25 on WebQA, 1.34 on TriviaQA, 0.49 on TruthfulQA, and 0.30 on HotpotQA.
The average increases from 37.65 to 39.28. These results indicate that the
benefit comes from conditionally modifying a strong direct-memory pathway rather
than simply combining all available representations.

After transferring the memory interface from Mistral to Llama, the resulting
system reaches an average of 37.87 and remains competitive with the
same-checkpoint Mistral E pathway. In particular, WebQA increases to 36.40.
Because the target backbone and adaptation history differ, this row shows that the memory interface remains usable after transfer rather than a matched gain over a bare Llama model.

\textbf{General NLP performance.}
Across the six NLP tasks, \textsc{MemoryAthena} improves over E by
3.90 points on SST2, 3.70 on MR, 1.70 on CR, 1.50 on RT, and 3.71 on AGN.
For these five tasks, we use the default admission threshold $\tau=0$.
For Yahoo, we use the more conservative setting $\tau=1$, under which the
router reaches 57.43 compared with 57.51 for E.
With these task-specific inference settings, the six-task average increases
from 76.73 to 79.13.

\takeaway{
The benefit of generated memory is heterogeneous across tasks and inputs:
neither GE nor GH uniformly dominates direct Engram retrieval. This is precisely
the regime targeted by \textsc{MemoryAthena}, which retains E as a stable
reference and selectively admits generated memories only when they are
predicted to help.
}
\subsection{RQ2: What drives the gains from the memory interface?}

We examine whether the gains arise from the memory interface itself, its
initialization, or the learned routing policy. We therefore compare
\textsc{MemoryAthena} with retrained architectural and addressing controls, a
from-scratch memory variant, and a random-router control. All trainable
configurations are evaluated over three random seeds, allowing us to assess
both average performance and run-to-run stability.
\begin{table}[!htbp]
\centering
\caption{
QA ablations across three random seeds.
Task scores are reported as mean $\pm$ standard deviation (\%).
Colored values denote absolute percentage-point changes relative to the
Vanilla Mistral no-memory baseline for individual tasks, while the colored
value in the Average column denotes the relative percentage change.
Average is the unweighted five-task mean, and the final column reports its
95\% confidence interval.
}
\label{tab:controls}
\small
\setlength{\tabcolsep}{3.2pt}
\resizebox{\linewidth}{!}{%
\begin{tabular}{lrrrrrrr}
\toprule
Condition
& NQ $\uparrow$
& WebQA $\uparrow$
& TriviaQA $\uparrow$
& TruthfulQA $\uparrow$
& HotpotQA $\uparrow$
& Average $\uparrow$
& 95\% CI \\
\midrule

Base (Vanilla Mistral, no memory)
& 20.18 $\pm$ 0.00
& 29.05 $\pm$ 0.00
& 57.55 $\pm$ 0.00
& \textbf{33.19} $\pm$ 0.00
& 20.48 $\pm$ 0.00
& 32.09 $\pm$ 0.00
& [32.09, 32.09] \\

\rowcolor{tablefocus}
Ours, pretrained memory
& \scorechange{\textbf{33.22} $\pm$ 0.31}{+13.04}
& \scorechange{\textbf{33.96} $\pm$ 0.90}{+4.90}
& \scorechange{\textbf{72.03} $\pm$ 1.23}{+14.48}
& \scorechange{31.83 $\pm$ 0.08}{-1.36}
& \scorechange{\textbf{27.06} $\pm$ 0.74}{+6.58}
& \avgchange{\textbf{39.62} $\pm$ 0.46}{+23.5}
& [38.47, 40.76] \\

\rowcolor{tablefocus}
Ours, from scratch
& \scorechange{31.97 $\pm$ 1.83}{+11.79}
& \scorechange{31.75 $\pm$ 2.20}{+2.69}
& \scorechange{69.27 $\pm$ 6.10}{+11.72}
& \scorechange{31.85 $\pm$ 0.23}{-1.34}
& \scorechange{26.28 $\pm$ 2.85}{+5.81}
& \avgchange{38.22 $\pm$ 2.45}{+19.1}
& [32.13, 44.32] \\

\groupbar{8}{Interface and addressing controls}

No gate
& \scorechange{30.70 $\pm$ 2.78}{+10.52}
& \scorechange{28.46 $\pm$ 5.08}{-0.59}
& \scorechange{65.17 $\pm$ 8.72}{+7.61}
& \scorechange{32.13 $\pm$ 0.14}{-1.06}
& \scorechange{24.61 $\pm$ 3.13}{+4.13}
& \avgchange{36.21 $\pm$ 2.78}{+12.8}
& [29.30, 43.13] \\

Permuted Engram keys
& \scorechange{27.50 $\pm$ 7.72}{+7.32}
& \scorechange{31.05 $\pm$ 0.62}{+1.99}
& \scorechange{59.58 $\pm$ 18.03}{+2.02}
& \scorechange{31.65 $\pm$ 0.88}{-1.54}
& \scorechange{22.61 $\pm$ 5.66}{+2.13}
& \avgchange{34.48 $\pm$ 6.44}{+7.4}
& [18.48, 50.47] \\

\groupbar{8}{Routing control}

Random router
& \scorechange{24.57 $\pm$ 0.52}{+4.39}
& \scorechange{32.20 $\pm$ 0.33}{+3.15}
& \scorechange{49.05 $\pm$ 0.14}{-8.50}
& \scorechange{31.69 $\pm$ 0.24}{-1.50}
& \scorechange{20.61 $\pm$ 0.06}{+0.13}
& \avgchange{31.63 $\pm$ 0.09}{-1.4}
& [31.40, 31.85] \\

\groupbar{8}{Alternative interfaces}

Parameter-matched FFN
& \scorechange{17.51 $\pm$ 4.18}{-2.67}
& \scorechange{26.20 $\pm$ 2.69}{-2.85}
& \scorechange{44.39 $\pm$ 6.20}{-13.16}
& \scorechange{30.76 $\pm$ 0.71}{-2.43}
& \scorechange{15.67 $\pm$ 2.45}{-4.81}
& \avgchange{26.91 $\pm$ 3.17}{-16.2}
& [19.03, 34.78] \\

Affine stitch
& \scorechange{31.81 $\pm$ 0.51}{+11.63}
& \scorechange{29.67 $\pm$ 2.61}{+0.62}
& \scorechange{70.29 $\pm$ 2.44}{+12.74}
& \scorechange{32.05 $\pm$ 0.47}{-1.14}
& \scorechange{27.00 $\pm$ 0.88}{+6.52}
& \avgchange{38.16 $\pm$ 0.57}{+18.9}
& [36.76, 39.57] \\

\bottomrule
\end{tabular}%
}
\end{table}
\textbf{Architectural and addressing controls.}
The pretrained-memory configuration achieves the highest five-task mean of
39.62, compared with 36.21 for the no-gate variant, 38.16 for affine stitch,
and 26.91 for the parameter-matched FFN. The full model outperforms both
no-gate and affine stitch on four of five tasks, and the parameter-matched FFN
on all five. Removing the gate reduces the average by 3.41 points, while
replacing the memory interface with a parameter-matched FFN reduces it by
12.71 points and even places performance below the no-memory baseline
(32.09). These results indicate that the improvement cannot be explained by
additional parameters alone and that the structured memory interface contributes
substantially to the observed gain.

The addressing control shows a similar pattern. Permuting the Engram keys
reduces the mean from 39.62 to 34.48. This condition is also substantially less
stable across seeds, with an average standard deviation of 6.44 compared with
0.46 for the pretrained-memory configuration. In particular, the large
variation on NQ, TriviaQA, and HotpotQA suggests that disrupting the association
between queries and stored Engrams can make memory use unreliable. Thus,
effective addressing appears to be an important component of the interface,
although the wide confidence interval of this control warrants caution in
interpreting the exact magnitude of the degradation.

\textbf{Memory initialization.}
Training the memory system from scratch remains effective, reaching a
five-task average of 38.22, well above the no-memory baseline of 32.09.
Pretrained memory is therefore not required for the interface to produce a
substantial gain. However, unlike the single-seed result, the three-seed
comparison favors pretrained initialization: it improves the mean from 38.22
to 39.62 and reduces the standard deviation from 2.45 to 0.46. The main role of
pretraining in this setting is therefore better characterized as improving
performance and cross-seed stability, rather than being a prerequisite for the
memory mechanism to work.

\textbf{Learned routing.}
The random-router control provides the clearest test of whether simply exposing
the model to multiple memory pathways is sufficient. Random routing reaches only
31.63 on average, compared with 39.62 for the learned router, a decrease of
7.99 points, and is slightly below the no-memory baseline of 32.09. The largest
task-level degradation occurs on TriviaQA, where performance drops from
72.03 to 49.05, while NQ falls from 33.22 to 24.57. Importantly, the random
router is highly stable across seeds ($31.63 \pm 0.09$), indicating that its
poor performance is not driven by an isolated failed run. Thus, access to E,
GE, and GH alone is insufficient: the model must learn when each memory
representation should intervene.

Because these experiments use three random seeds, several high-variance
controls have wide confidence intervals. We therefore interpret the results
primarily in terms of consistent effect size and stability rather than formal
statistical significance. Complete task-level ablation results are reported in
Appendix~\ref{app:ablations}.

\takeaway{
The gains are driven by the combination of a structured memory interface,
reliable addressing, and learned routing. The pretrained-memory configuration
achieves the highest average performance (39.62) with low cross-seed variation,
whereas removing the gate, permuting memory keys, or replacing the interface
with a parameter-matched FFN degrades performance. Random routing removes
essentially the entire gain, falling to 31.63, showing that multiple memory
pathways are useful only when their use is selectively learned. Training from
scratch still performs strongly (38.22), indicating that pretrained memory is
not necessary for the mechanism to work, but it improves both average
performance and stability.
}
\subsection{RQ3: How does \textsc{MemoryAthena} use complementary memory pathways?}
\label{sec:rq3}
\begin{figure}[!htbp]
    \centering

    \begin{subfigure}[t]{0.50\columnwidth}
        \centering
        \safeincludegraphics[width=\linewidth]{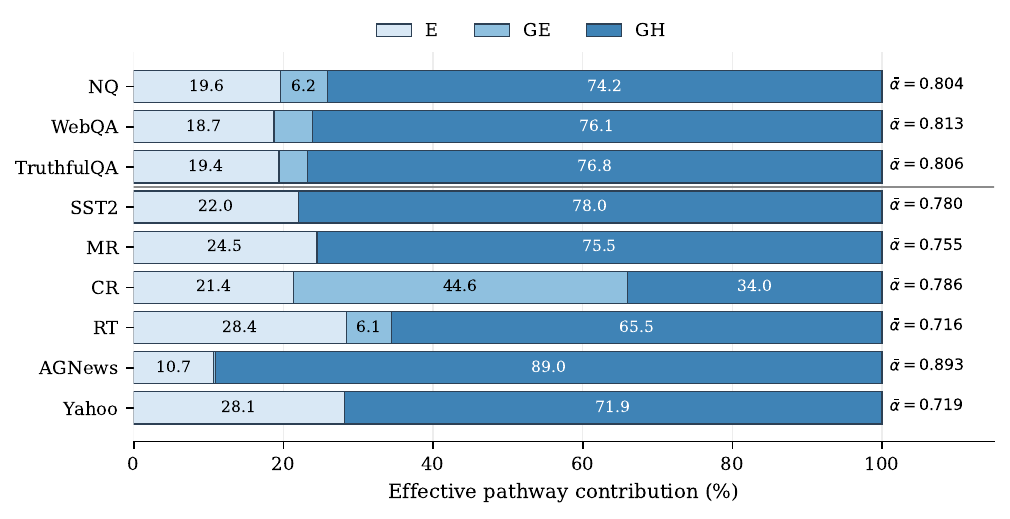}
        \caption{Downstream memory-pathway contribution.}
        \label{fig:downstream-routing}
    \end{subfigure}
    \hfill
    \begin{subfigure}[t]{0.46\columnwidth}
        \centering
        \safeincludegraphics[width=\linewidth]{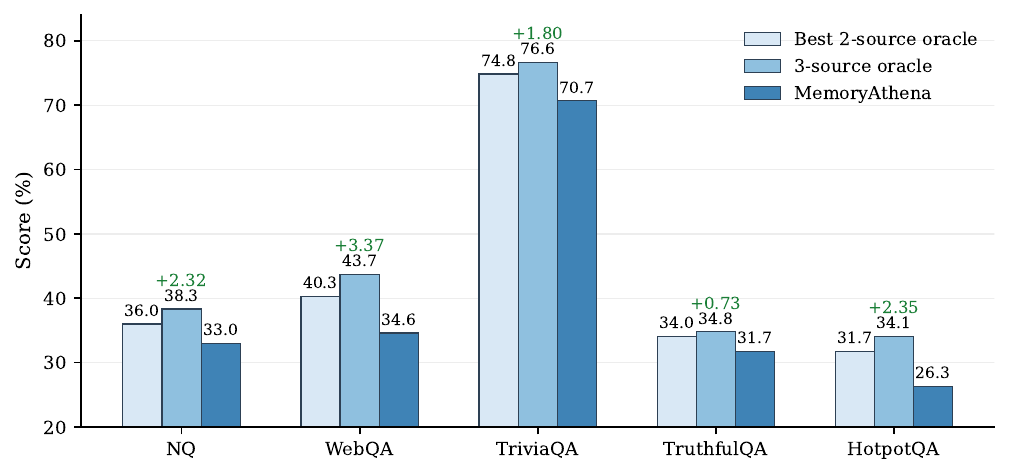}
        \caption{Gold-label oracle headroom.}
        \label{fig:oracle-headroom}
    \end{subfigure}

    \caption{
    \textbf{Left:} effective downstream contribution of E, GE, and GH across tasks.
    \textbf{Right:} comparison between deployed \textsc{MemoryAthena} and
    gold-label source oracles, showing the remaining routing headroom.
    }
    \label{fig:routing-and-oracle}
\end{figure}

The previous results establish that generated memory can improve downstream
performance, but not how the router combines the three pathways. We therefore analyze the downstream routing behavior and,
separately, use a label-informed oracle to estimate how much complementary
information remains unexploited.

\textbf{Downstream routing behavior.}
When a generated pathway $s^\star$ is admitted, the effective memory
contribution can be written as
\begin{equation}
(w_{\mathrm{E}},w_{\mathrm{GE}},w_{\mathrm{GH}})
=
\left(
1-\alpha,\;
\alpha\mathbf{1}[s^\star=\mathrm{GE}],\;
\alpha\mathbf{1}[s^\star=\mathrm{GH}]
\right).
\end{equation}
We aggregate these effective weights over downstream inference to characterize
how much each pathway contributes to the injected memory representation.

Figure~\ref{fig:downstream-routing} shows task-dependent routing.
On the available QA traces, GH provides the majority of the memory contribution,
74.15\% on NQ, 76.08\% on WebQA, and 76.77\% on TruthfulQA, whereas GE
contributes 3.83--6.23\%. A similar preference for GH appears on several
classification tasks, including SST2 (77.99\%), MR (75.50\%), AGN (89.03\%), and Yahoo (71.86\%).

CR is the counterexample. There GE receives 44.60\% of the contribution,
against 34.04\% for GH and 21.36\% for E, and RT also retains a larger
direct-memory component (28.42\%). The mean interpolation strength ranges from $0.716$ to $0.893$. An admitted
generated memory therefore typically makes a large correction to the direct
representation.
These percentages are effective interpolation weights. Because E retains the
$(1-\alpha)$ component whenever GE or GH is admitted, a source's contribution
differs from the fraction of tokens routed exclusively to it.

\textbf{Oracle headroom.}
The routing statistics describe what the deployed router does, whereas a
post-hoc source oracle estimates the gain that was attainable. It observes the downstream label
and selects the highest-scoring endpoint among the available pathways.
The three-source oracle improves over the best E-containing two-source oracle
on every task, with gains of 2.32 points on NQ, 3.37 on WebQA, 1.79 on
TriviaQA, 0.73 on TruthfulQA, and 2.35 on HotpotQA
(Figure~\ref{fig:oracle-headroom}). This means that GE and GH provide successful
predictions on examples that are not captured by a single generated pathway.
A gap remains between the label-informed oracle and the deployed causal
router. The three-source oracle exceeds
\textsc{MemoryAthena} by 5.31 points on NQ, 9.07 on WebQA, 5.94 on TriviaQA,
3.04 on TruthfulQA, and 7.74 on HotpotQA. Diverse memory representations are already available, and much of their
potential therefore depends on identifying \emph{when} a generated memory is
useful and \emph{which} generated pathway should intervene.

\takeaway{
The utility of E, GE, and GH is strongly task-dependent, confirming that no
single memory pathway is uniformly preferable. Gold-label oracles exceed the
deployed router by 3.04--9.07 points, showing that substantial complementarity
among the pathways remains unexploited. The key challenge is therefore to
better identify when a generated memory should intervene and which pathway is
most useful.
}
\section{Conclusion}

\textsc{MemoryAthena} starts from the observation that generated memory is
conditionally useful rather than uniformly superior to direct retrieval.
Instead of replacing E, the method treats GE and GH as candidate corrections
and learns when and how strongly they should intervene. This converts
heterogeneous pathway quality into consistent gains on the QA summary metrics
and improvements across the evaluated NLP setting. The remaining oracle gap
suggests that better utility estimation and admission, rather than universally
stronger generated memories, is a central direction for future work. Future work we will also explore how to use agentic RL methods to optimize the memory~\cite{xie2026step}.

\subsubsection*{Reproducibility statement}
The appendices specify every pathway, freezing boundary, objective, admission rule and control, together with sample counts and checkpoint selection, and the accompanying evidence ledger maps each table to its artifacts. Available hashes support traceability but do not substitute for an immutable launch environment or contamination audit. Missing uncertainty estimates or functional coding scores are not inferred.

\subsubsection*{AI assistance}
An AI assistant assisted drafting, organization, and consistency checks against experiment records. The claims, experimental validity, and final submission remain the authors' responsibility.

\bibliography{memoryathena_references}
\clearpage
\appendix

\section{Evaluation datasets and protocol}\label{app:setup}

The primary backbone is Mistral-7B-v0.3 \citep{jiang2023mistral}. The QA router uses Wikipedia-2021 text and imports a learned Llama-2 source memory \citep{touvron2023llama}. General NLP uses an equal-token mixture of WikiText-103 \citep{merity2017pointer}, Amazon Polarity review text \citep{zhang2015character}, CC-News \citep{blagojevic2021ccnews}, and IMDB text \citep{maas2011learning}. All parameters are frozen downstream.

QA covers Natural Questions (NQ) \citep{kwiatkowski2019natural}, WebQuestions (WebQA) \citep{berant2013semantic}, TriviaQA \citep{joshi2017trivia}, TruthfulQA \citep{lin2022truthful}, and HotpotQA \citep{yang2018hotpot}. We report exact match (EM) and token F1 for open QA, and the arithmetic mean of three recorded multiple-choice metrics for TruthfulQA. Sample counts are 3,609, 2,032, 17,944, 817, and 7,405. Exact metrics are retained in Appendix~\ref{app:qa}.

For general NLP, SST2 \citep{socher2013recursive}, MR and RT (movie-review sentiment benchmarks, see \citealp{pang2005seeing}), and CR \citep{hu2004mining} test sentiment, while AG News (AGN) and Yahoo Answers test topic classification \citep{zhang2015character}. SST2 uses its development split, MR and CR use the provided test files, and RT uses the Rotten Tomatoes test split. The exact release identifiers and preprocessing lineage of these MR/RT files remain to be verified, and the citation identifies the benchmark family. Our next-token scoring adapts domain-conditional pointwise mutual information \citep{holtzman2021surface}. For label-token set $V_y$,
\begin{equation}
 S(y;x)=\sum_{v\in V_y}
       [\log p(v\mid C(x))-\log p(v\mid C_{\mathrm{domain}})],
 \qquad \widehat y=\arg\max_y S(y;x).
 \label{eq:dcpmi}
\end{equation}
Verbalizers map to their first valid token and are deduplicated following the implementation. This sums log-score differences token by token rather than forming a log-sum-exp or a full-label likelihood. The six tasks contain 872, 2,000, 2,000, 1,066, 7,600, and 60,000 examples. We report accuracy and an unweighted six-task mean.

Our primary comparison keeps the memory and expert checkpoints fixed and changes
only the inference rule, providing the cleanest assessment of routing.
Retrained ablations are reported separately because they also change the learned
components. The routing models are trained on general text without downstream
labels.

\input{appendix}
\end{document}

%% file: appendix.tex
\section{Limitations and Future Work}

Although \textsc{MemoryAthena} learns when to admit GE or GH from unlabeled
text, the overall routing system still relies on memory
pathways, routing features, training objectives, and admission hyperparameters.
It therefore does not yet provide a fully automatic mechanism for discovering
which memory representation should be constructed and used for a given input.
The substantial gap between the deployed router and the gold-label source
oracle further shows that the available memory pathways contain useful
complementary information that the current router does not fully exploit.
Developing more automated routing objectives that can jointly discover useful
memory candidates, calibrate their utility, and approach oracle-level selection
without downstream labels is an important direction for future work.

\section{Architecture and routing implementation}\label{app:implementation}

\subsection{Direct and generated pathways}

E directly reads retrieved memory. GE generates a small latent representation from a causal window of Engram cues. GH generates from clean causal backbone states, obtained without memory injection:
\begin{align}
 e_t^\ell &=R_E^\ell(h_t^\ell,m_t), \nonumber\\
 z_{E,t}^\ell &=G^\ell(M_{t-w+1:t};\E),&
 z_{H,t}^\ell &=G^\ell(H^{0,\ell}_{t-w+1:t};H), \label{eq:sources}\\
 g_{\mathrm{GE},t}^\ell &=R_{GE}^\ell(h_t^\ell,z_{E,t}^\ell),&
 g_{\mathrm{GH},t}^\ell &=R_{GH}^\ell(h_t^\ell,z_{H,t}^\ell).\nonumber
\end{align}
Reader notation subsumes projections, branch aggregation, and output gates. Generated paths share components and use source embeddings and low-rank adaptations. GE is grounded in retrieved cues, whereas GH can propose a representation when the table is unhelpful. This creates potential complementarity, but also makes strong performance with corrupted memory possible.

The evaluated Mistral configuration injects memory at layers 2 and 10. Memory dimension is 512 and the reader has four branches. Each generator produces four latents from a three-position causal window, with width 256, two layers, and four attention heads. Source-adaptation rank is 16. GH requires a clean backbone computation in the current implementation, and freezing its parameters does not remove this inference cost.

\subsection{Memory learning and reader adaptation}

Training comprises three optimization stages followed by frozen evaluation. Stage 1 learns the table and source adaptor with the source backbone fixed:
\begin{equation}
 \mathcal L_{\mathrm{mem}}(\phi,\psi_{\mathrm{src}})
 =-\sum_t\log p_{\theta_{\mathrm{src}},\phi,\psi_{\mathrm{src}}}
 (x_{t+1}\mid x_{\leq t}).
 \label{eq:memloss}
\end{equation}
The table and canonicalization configuration can be reused, but the source adaptor need not be compatible with another backbone.

Stage 2 freezes the table and target backbone and trains the generators and target-side readers. In the later general-NLP and coding pipeline, each endpoint receives equal-weight causal language-modeling supervision:
\begin{equation}
 \mathcal L_{\mathrm{experts}}
 =-\frac13\sum_{s\in\{\E,\GE,\GH\}}\sum_t
 \log p_s(x_{t+1}\mid x_{\leq t}).
 \label{eq:expertloss}
\end{equation}
This makes the endpoints usable before their relative advantages are distilled. It does not optimize downstream labels.

The QA checkpoint also contains auxiliary routing modules used to construct the
baseline inference rules in Table~\ref{tab:qa_combined}.
\emph{Ordinary hard routing} treats E, GE, and GH as three symmetric candidates
and selects the single pathway with the highest routing score.
\emph{Ordinary soft fusion} instead applies learned softmax weights over all
three pathways and combines their residuals continuously.
The \emph{subset} variants restrict routing to learned subsets of the available
pathways. \emph{Subset hard routing} makes a discrete choice within the selected
subset, whereas \emph{subset soft fusion} uses the learned mixture weights
within that subset.
These auxiliary routers are trained as part of the QA reader checkpoint and are
used only as comparison baselines. In contrast, the final
\textsc{MemoryAthena} router operates over E, GE, and GH using E-relative
advantage prediction: E is kept as the anchor, and GE or GH modifies it only
when the generated pathway is predicted to be beneficial.

\subsection{Counterfactual future-advantage supervision}

Stage 3 freezes the backbone, memory, generators, and readers. Under teacher forcing, the advantage of source $s$ over E is
\begin{equation}
 a_{s,t}=\log p_s(x_{t+1}\mid x_{\leq t})
       -\log p_E(x_{t+1}\mid x_{\leq t}),\qquad a_{E,t}=0.
 \label{eq:advantage}
\end{equation}
A positive advantage corresponds to a lower next-token loss than E. Each endpoint pass forces the same source at all injection sites. The advantage is therefore a full-path counterfactual and cannot be attributed to one layer.

We average over future horizons $\mathcal H=\{1,4,8,16,32\}$. Let $v_j$ indicate a valid target and $\mathcal H_t$ contain horizons with at least one valid target:
\begin{equation}
 A_{s,t}=\frac1{|\mathcal H_t|}\sum_{k\in\mathcal H_t}
 \frac{\sum_{j=0}^{k-1}v_{t+j}a_{s,t+j}}
      {\sum_{j=0}^{k-1}v_{t+j}} .
 \label{eq:horizons}
\end{equation}
Out-of-range targets are excluded. Future text constructs offline supervision only, so the deployed head sees neither future tokens nor these targets.

At each injection layer, the routing head predicts $\widehat A_{s,t}^\ell$ and confidence logit $c_{s,t}^\ell$ from detached causal features. Features include the hidden state, reader-gate strength, residual magnitudes, pairwise similarities, active-source indicators, and projected semantic features. E's predicted advantage is fixed at zero. The head preserves the E stream while exposing generated alternatives as detached features.

For $s\in\{\GE,\GH\}$, define $u_{s,t}=\clip(|A_{s,t}|,0.01,2)$. The per-position loss is
\begin{align}
 L_{s,t}^{\ell}
 &=\operatorname{SmoothL1}\left(\widehat A_{s,t}^\ell,
                         \clip(A_{s,t},-2,2)\right) \nonumber\\
 &\quad+0.25\,\operatorname{BCEWithLogits}
            \left(c_{s,t}^\ell,\sigma(A_{s,t}/0.15)\right).
 \label{eq:headloss}
\end{align}
We take a valid-position, $u_{s,t}$-weighted average and then average injection-layer heads. Only these heads are optimized, amounting to 534,924 trainable parameters in the audited Mistral runs. Confidence is trained against an advantage-derived soft target, and it is not established to be a calibrated probability of downstream correctness.
\subsection{E-Anchored Causal Inference}

At each position and injection layer, a generated pathway
$s\in\{\mathrm{GE},\mathrm{GH}\}$ is considered only if its predicted
E-relative advantage exceeds $\tau$ and its confidence exceeds $\rho$.
If both generated pathways are eligible, we select the one with the larger
predicted advantage.
For the selected pathway $s^\star$, the intervention strength is
\begin{equation}
\alpha_t^\ell =
a_{\max}
\operatorname{clip}
\left(
\frac{\widehat A_{s^\star,t}^\ell-\tau}{T_\alpha},
0,1
\right)
\sigma(c_{s^\star,t}^\ell),
\end{equation}
and the memory residual is
\begin{equation}
r_t^\ell
=
(1-\alpha_t^\ell)e_t^\ell
+
\alpha_t^\ell g_{s^\star,t}^\ell .
\end{equation}

If no generated pathway is admitted, $\alpha_t^\ell=0$ and the model exactly
recovers the E residual at that injection site. We use
$\tau=0$, $\rho=0.5$, $T_\alpha=0.15$, and $a_{\max}=1$ by default.

The fallback guarantees recovery of the same-checkpoint E pathway when all
generated candidates are rejected, but it does not guarantee that every
admitted intervention improves downstream performance because the router can
make incorrect predictions.
For analysis, the corresponding effective pathway weights are
\begin{equation}
(w_{\mathrm E},w_{\mathrm{GE}},w_{\mathrm{GH}})
=
\left(
1-\alpha,\,
\alpha\mathbf{1}[s^\star=\mathrm{GE}],\,
\alpha\mathbf{1}[s^\star=\mathrm{GH}]
\right).
\end{equation}
These weights measure contribution to the injected residual and should not be
confused with discrete source-selection frequencies.
\section{Training Configuration and Model Architecture}
\label{app:training}

\subsection{Training Configuration}

We use a common budget of 20M processed input positions for each optimization
stage. The three stages separately learn the memory, adapt the generated-memory
interfaces, and train the final routing head. During downstream evaluation, all
model parameters are frozen.

\begin{table}[t]
\centering
\caption{
Training stages and token budgets. Each optimization stage uses at most
20M processed input positions.
}
\label{tab:tokenbudget}
\small
\setlength{\tabcolsep}{5pt}
\begin{tabularx}{\linewidth}{@{}lXr@{}}
\toprule
\textbf{Stage} & \textbf{Trainable components} & \textbf{Budget} \\
\midrule
Memory learning
& Engram memory table and source-side adaptor
& 20M \\

Memory-interface adaptation
& GE/GH generators and target-side readers
& 20M \\

Router training
& E-relative advantage and confidence heads
& 20M \\

Evaluation
& None; all components are frozen
& -- \\
\bottomrule
\end{tabularx}
\end{table}

For the QA experiments, the learned source memory is reused rather than
retrained during the final routing run. Thus, the 20M budget in
Table~\ref{tab:tokenbudget} describes the budget of each optimization stage in
the full pipeline, not an additional 20M-token memory-training phase for every
downstream experiment.

\begin{table}[t]
\centering
\caption{
Router training and inference configuration for the evaluated
Mistral-7B-v0.3 models.
}
\label{tab:routerconfig}
\small
\setlength{\tabcolsep}{6pt}
\begin{tabularx}{\linewidth}{@{}Xr@{}}
\toprule
\textbf{Setting} & \textbf{Value} \\
\midrule
Packed sequence length & 2,048 \\
Training steps & 9,765 \\
Processed input positions & 19,998,720 \\
Validation budget & 2,000,000 positions \\
Trainable router parameters & 534,924 \\
Router hidden width & 64 \\
Learning rate & $3\times10^{-4}$ \\
Batch size & 1 \\
Warmup steps & 200 \\
Validation interval & 2,000 steps \\
\addlinespace
QA selected checkpoint
& step 4,000 (8.192M positions) \\
General-NLP selected checkpoint
& step 8,000 (16.384M positions) \\
\addlinespace
Advantage threshold $\tau$ & 0.0 \\
Confidence threshold $\rho$ & 0.5 \\
Maximum interpolation $a_{\max}$ & 1.0 \\
Interpolation temperature $T_{\alpha}$ & 0.15 \\
\bottomrule
\end{tabularx}
\end{table}

The final checkpoint does not have to coincide with the end of training.
Router checkpoints are selected using held-out causal-text validation rather
than downstream task accuracy. The QA router selects the checkpoint at
8.192M processed positions, while the general-NLP router selects the checkpoint
at 16.384M positions. The separate Yahoo threshold adjustment is reported
explicitly because it uses downstream test performance for model selection.

\subsection{Architecture Specification}

Table~\ref{tab:architecture} summarizes the memory interface, generated-memory
modules, and routing head used with Mistral-7B-v0.3.

\begin{table}[t]
\centering
\caption{
Architecture and parameterization of the generated-memory pathways and
E-anchored router used with Mistral-7B-v0.3.
}
\label{tab:architecture}
\small
\setlength{\tabcolsep}{5pt}
\renewcommand{\arraystretch}{1.08}

\begin{tabularx}{\linewidth}{
    @{}
    p{0.30\linewidth}
    X
    >{\raggedleft\arraybackslash}p{0.24\linewidth}
    @{}
}
\toprule
\textbf{Component} & \textbf{Configuration} & \textbf{Value} \\
\midrule

\multicolumn{3}{@{}l}{\textbf{Memory interface}} \\
\addlinespace[2pt]

Injection layers
& Target backbone layers
& $\{2,10\}$ \\

Memory dimension
& Retrieved / latent memory width
& 512 \\

Memory table
& Frozen Engram parameters
& 33,554,432 \\

Reader branches
& Parallel branches per injection layer
& 4 \\

Direct E reader
& Parameters per layer / two layers
& 10,506,244 / 21,012,488 \\

\addlinespace[3pt]
\midrule
\multicolumn{3}{@{}l}{\textbf{Generated-memory pathways (GE/GH)}} \\
\addlinespace[2pt]

Generator context
& Causal input window
& 3 positions \\

Generated latents
& Latent representations per position
& 4 \\

Generator hidden width
& Internal representation size
& 256 \\

Generator depth
& Transformer layers
& 2 \\

Generator attention
& Attention heads
& 4 \\

Source-specific adaptation
& Low-rank output adapter
& $r=16$ \\

Output-adapter shape
& Bottleneck projection
& $256 \rightarrow 16 \rightarrow d_{\mathrm{model}}$ \\

GE conditioning
& Retrieved Engram cues
& -- \\

GH conditioning
& Clean causal backbone states
& -- \\

Generated reader
& Projection, branch aggregation, and output gating
& -- \\

\addlinespace[3pt]
\midrule
\multicolumn{3}{@{}l}{\textbf{E-anchored router}} \\
\addlinespace[2pt]

Router architecture
& MLP hidden width
& 64 \\

Candidate pathways
& Generated candidates
& $\{\mathrm{GE},\mathrm{GH}\}$ \\

Reference pathway
& Fixed routing anchor
& E \\

Router outputs
& E-relative advantage and confidence
& 2 per candidate \\

Advantage threshold
& $\tau$
& 0.0 \\

Confidence threshold
& $\rho$
& 0.5 \\

Maximum interpolation
& $a_{\max}$
& 1.0 \\

Interpolation temperature
& $T_{\alpha}$
& 0.15 \\

\addlinespace[3pt]
\midrule
\multicolumn{3}{@{}l}{\textbf{Parameter summary}} \\
\addlinespace[2pt]

Shared generator
& Parameters per layer / two layers
& 3,288,576 / 6,577,152 \\

Generated readers
& Parameters per layer / two layers
& 69,625,352 / 139,250,704 \\

Advantage router
& Parameters per layer / two layers
& 267,462 / 534,924 \\

Full tri-path adaptor
& All adaptor and routing parameters per layer
& 83,960,284 \\

Two-layer adaptor
& Layers 2 and 10, excluding backbone
& 167,920,568 \\
\addlinespace[2pt]
\textbf{Total memory system}
& \textbf{Memory table + two-layer adaptor, excluding backbone}
& \textbf{201,475,000} \\

\bottomrule
\end{tabularx}
\end{table}
\paragraph{Reader and generator training.}
Stage 2 freezes the memory table and target backbone and optimizes the
generated-memory modules and target-side readers. For general NLP and coding,
the three endpoints E, GE, and GH are trained with the equal-weight
language-modeling objective in Eq.~\ref{eq:expertloss}. This stage contains
167,364,632 trainable parameters.

The QA reader checkpoint uses the same three endpoints but additionally trains
auxiliary pair-restricted and subset-fusion modules used by the hard- and
soft-routing baselines. These auxiliary modules are not additional memory
sources in the final system: the deployed \textsc{MemoryAthena} router operates
only over E, GE, and GH. The QA reader/generator stage contains 167,385,644
trainable parameters and uses routing distillation with coefficient 0.5.

\paragraph{Training time.}
The recorded reader/generator training times are approximately 31.70 hours for
QA, 16.01 hours for general NLP, and 16.41 hours for coding. Router training
takes approximately 11.19, 10.58, and 10.94 hours, respectively. These values
are wall-clock measurements from the corresponding runs. Because router
training evaluates multiple counterfactual endpoints, and GH additionally uses
a clean-backbone forward pass, processed-token count alone does not represent
the total computational cost.

\paragraph{General-NLP training data.}
The general-NLP training corpus uses an equal-token mixture of WikiText-103,
Amazon Polarity, CC-News, and IMDB. No downstream labels are used for training
the memory pathways or the router. We nevertheless do not assume that this
guarantees benchmark decontamination: review corpora may overlap with
downstream review benchmarks, and the CC-News training and validation streams
are sampled from the same underlying split rather than from explicitly
document-disjoint partitions.
\section{Complete QA Metrics}
\label{app:qa}

\begin{figure}[t]
    \centering
    \includegraphics[width=\columnwidth]{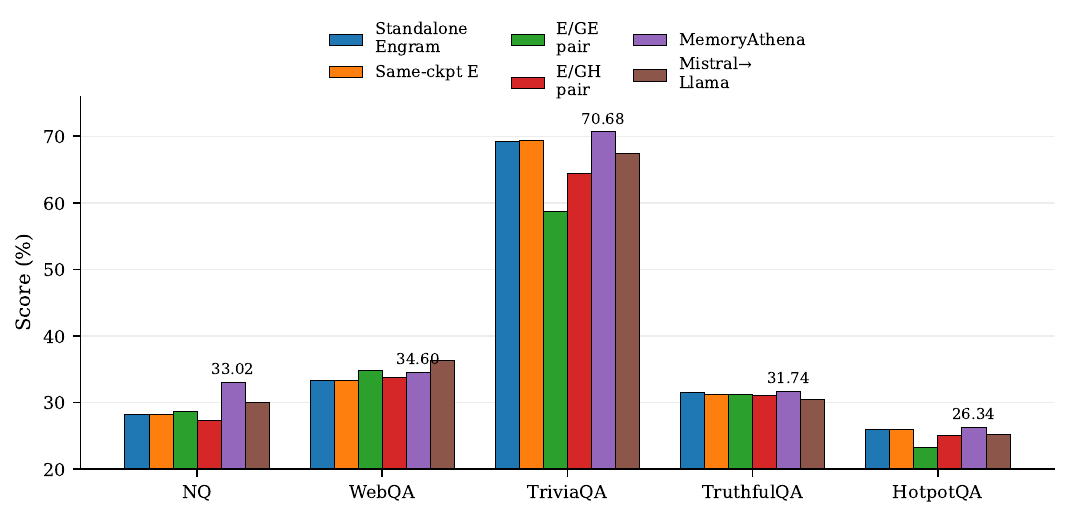}
    \caption{
    Complete QA comparison. Open-QA tasks report F1, and TruthfulQA reports the
    mean of MC1, MC2, and MC3. Same-checkpoint rows isolate the effect of the
    inference rule more cleanly than Standalone Engram or
    Mistral$\rightarrow$Llama.
    }
    \label{fig:qa-summary}
\end{figure}

Figure~\ref{fig:qa-summary} summarizes the complete QA comparison using F1 for
open-QA tasks and the mean multiple-choice score for TruthfulQA. The
same-checkpoint rows provide the cleanest comparison because they share the same
memory and expert checkpoints. Standalone Engram and Mistral$\rightarrow$Llama
serve as additional references but have different training histories.

The E+GE and E+GH pair controls are learned two-source fusion baselines.
Their mixing weights depend on the input, so they should be interpreted as
complete learned-fusion systems rather than as isolated measurements of the
contribution of GE or GH alone.

\begin{figure}[t]
    \centering
    \includegraphics[width=\columnwidth]{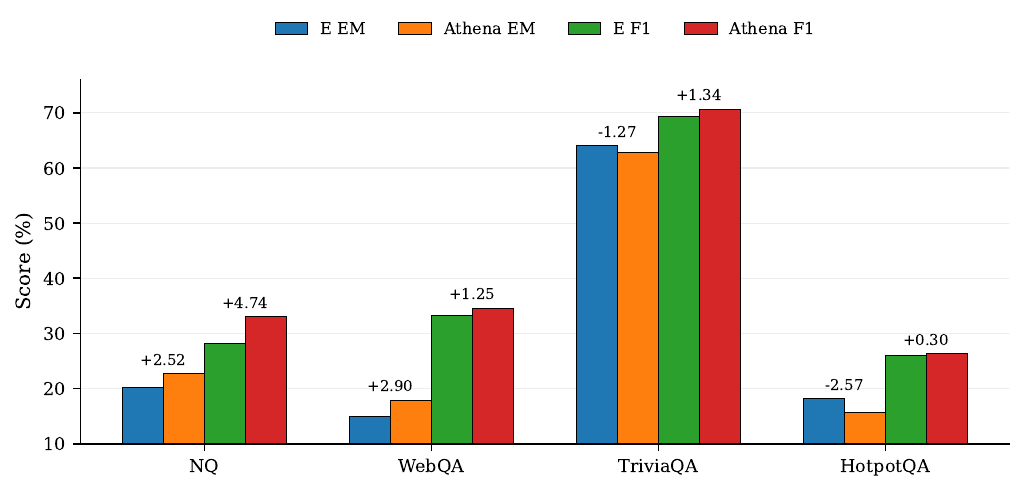}
    \caption{
    EM/F1 comparison between same-checkpoint E and \textsc{MemoryAthena} on the
    open-QA tasks. Numbers above bars show the change from E to
    \textsc{MemoryAthena}. TriviaQA and HotpotQA exhibit higher F1 but lower EM.
    }
    \label{fig:qa-emf1}
\end{figure}

To make the answer-level behavior explicit, Figure~\ref{fig:qa-emf1} compares
EM and F1 between same-checkpoint E and \textsc{MemoryAthena}. NQ and WebQA
improve on both metrics, but TriviaQA and HotpotQA show a different pattern:
F1 increases while EM decreases. Thus, generated-memory routing can improve
average answer overlap without necessarily increasing exact-match accuracy.

For TruthfulQA, Table~\ref{tab:truthfulqa-mc} reports the three multiple-choice
metrics separately.

\begin{table}[h]
\centering
\caption{
TruthfulQA multiple-choice metrics (\%). The summary is the mean of MC1, MC2,
and MC3.
}
\label{tab:truthfulqa-mc}
\small
\begin{tabular}{lrrrr}
\toprule
Condition & MC1 & MC2 & MC3 & Mean\\
\midrule
Standalone Engram & 27.42 & 44.32 & 22.69 & 31.47\\
Same-checkpoint E & 26.93 & 44.18 & 22.64 & 31.25\\
E/GE learned pair & 27.54 & 43.40 & 22.72 & 31.22\\
E/GH learned pair & 27.78 & 42.90 & 22.67 & 31.12\\
\rowcolor{tablefocus}
\textsc{MemoryAthena} & 28.15 & 44.04 & 23.02 & 31.74\\
Mistral$\rightarrow$Llama & 27.78 & 41.57 & 21.87 & 30.41\\
\bottomrule
\end{tabular}
\end{table}

\begin{figure}[t]
    \centering
    \includegraphics[width=\columnwidth]{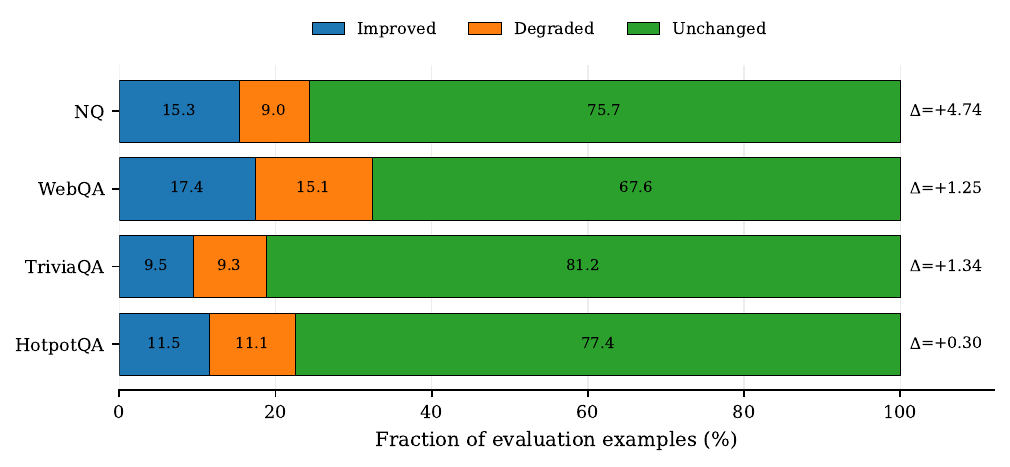}
    \caption{
    Paired F1 changes relative to same-checkpoint E. Bars show the fraction of
    evaluation examples whose F1 improves, degrades, or remains unchanged under
    \textsc{MemoryAthena}. The right margin reports the mean F1 change.
    These counts are descriptive only and are not significance tests.
    }
    \label{fig:qa-paired}
\end{figure}

Finally, Figure~\ref{fig:qa-paired} reports paired F1 changes relative to
same-checkpoint E. Most examples remain unchanged, while the numbers of
improved and degraded examples are of similar order. Positive mean F1 gains
therefore arise from the magnitudes of the changes rather than from uniformly
one-sided answer flips.

Paired improve/degrade counts alone do not determine the mean score change,
because the sizes of the answer-level changes can differ substantially.
We therefore report these counts as descriptive diagnostics only. TruthfulQA is
excluded from this paired F1 analysis because it does not use token-overlap F1.
\section{Source-Oracle Analysis}
\label{app:oracle}

Figure~\ref{fig:oracle-headroom} evaluates the complementarity of the three
memory pathways using a gold-label oracle. For each example, we run the E, GE,
and GH endpoints separately and select the endpoint that obtains the highest
score against the gold answer. This oracle is used only for analysis and is
not available at inference time.

Adding more candidate endpoints cannot reduce oracle performance, since the
oracle can always retain the best previously available endpoint. Therefore,
the improvement from the two-source to the three-source oracle indicates that
the additional pathway is useful on some examples, but does not imply that a
deployable router can always identify those examples.

\begin{table}[h]
\centering
\caption{
Number and percentage of examples for which each endpoint is selected by the
gold-label oracle. Ties are resolved in the order E, GE, GH.
}
\begin{tabular}{lrrr}
\toprule
Task & E & GE & GH\\
\midrule
NQ &2,854 (79.08)&475 (13.16)&280 (7.76)\\
WebQA &1,428 (70.28)&336 (16.54)&268 (13.19)\\
TriviaQA &15,764 (87.85)&1,345 (7.50)&835 (4.65)\\
TruthfulQA &361 (44.19)&254 (31.09)&202 (24.72)\\
HotpotQA &6,150 (83.05)&762 (10.29)&493 (6.66)\\
\bottomrule
\end{tabular}
\end{table}

E is the most frequent oracle winner, although its count is increased by our
tie-breaking rule: examples on which multiple endpoints receive the same score
are assigned to E first. Importantly, these counts describe which complete
endpoint performs best on each example; they are not routing frequencies of
\textsc{MemoryAthena}. The deployed router mixes pathways locally across
tokens and layers, so its average pathway weights measure a different quantity. This does not contradict the GH-dominated average routing weights reported
elsewhere: the two statistics measure different quantities. Oracle counts
identify the best complete endpoint per example, whereas routing weights measure
the local contribution of each pathway within the deployed mixed trajectory.
\section{Ablations and Architectural Controls}
\label{app:ablations}

\begin{figure}[h]
    \centering
    \includegraphics[width=\columnwidth]
    {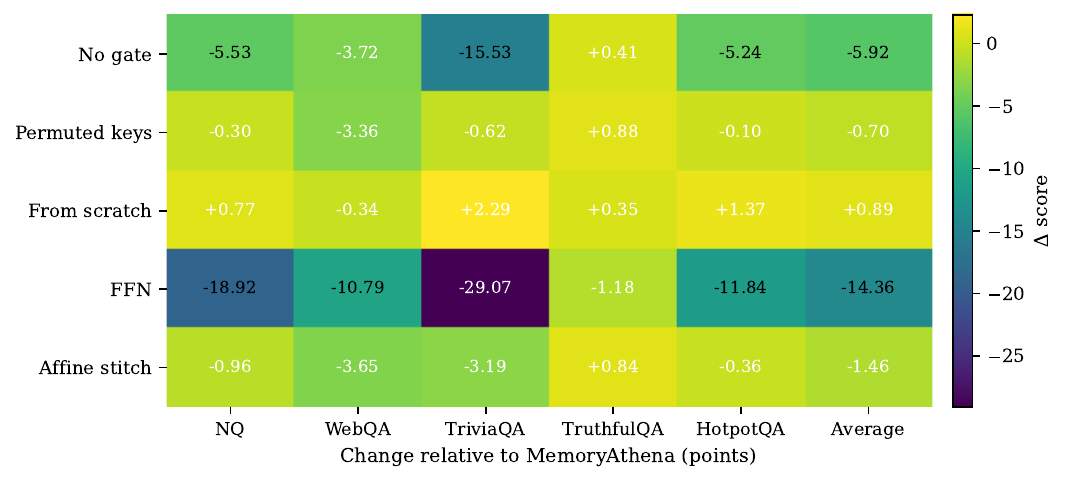}
    \caption{
    Performance changes relative to the full \textsc{MemoryAthena} system.
    Open-QA tasks use F1 and TruthfulQA uses the mean multiple-choice score.
    The final column reports the change in the five-task average.
    }
    \label{fig:ablation-delta}
\end{figure}

Figure~\ref{fig:ablation-delta} summarizes the main architectural ablations
relative to \textsc{MemoryAthena}. For open-QA tasks, we report F1; for
TruthfulQA, we report the mean of the three multiple-choice metrics. Positive
values indicate an improvement over the full system, and negative values
indicate a degradation.

Removing the output gate causes the largest degradation among the
memory-interface ablations, reducing the five-task average by 5.92 points.
Replacing the learned interface with a parameter-matched FFN is substantially
weaker, with an average drop of 14.36 points. The affine-stitch variant also
underperforms the full system by 1.46 points on average.

Permuting the memory keys has a comparatively small effect on the aggregate
score, decreasing the five-task average by 0.70 points. In contrast, training
the memory system from scratch reaches a slightly higher average than the
pretrained-memory configuration (+0.89 points), with improvements on NQ,
TriviaQA, TruthfulQA, and HotpotQA but a small decrease on WebQA.

These results suggest that the learned memory interface and its gating
mechanism are important for performance, whereas the advantage of the
pretrained memory initialization is less consistent across the evaluated QA
tasks.
\section{Threshold Sensitivity}
\label{app:classification}

We study the sensitivity of the router to the advantage threshold $\tau$ on
Yahoo Answers by varying $\tau$ while keeping the remaining inference
configuration fixed. As shown in Table~\ref{tab:yahoo-threshold}, performance
improves steadily as the threshold becomes more conservative, with the largest
changes occurring between $\tau=0.3$ and $\tau=0.9$. Performance then largely
stabilizes around $\tau=0.9$--$1.0$, indicating that routing behavior is
sensitive to the admission threshold but becomes relatively stable in the
high-threshold regime.

\begin{table}[h]
\centering
\caption{
Sensitivity of Yahoo Answers accuracy to the routing advantage threshold
$\tau$ ($N=60{,}000$).
}
\label{tab:yahoo-threshold}
\begin{tabular}{lr}
\toprule
Threshold $\tau$ & Accuracy (\%)\\
\midrule
0.00 & 45.8967\\
0.05 & 46.1150\\
0.10 & 46.2483\\
0.20 & 47.0283\\
0.30 & 48.3583\\
0.90 & 57.3600\\
1.00 & 57.4317\\
\bottomrule
\end{tabular}
\end{table}

These results show that the admission threshold can have a substantial effect
on downstream classification performance. A larger $\tau$ makes the router
more selective about when generated memory is allowed to intervene, suggesting
that conservative routing is particularly important on Yahoo Answers.

As an additional classification result, the router reaches 72.56\% accuracy on
RTE, compared with 71.48\% for the E pathway, over 277 evaluation examples.
\section{Interpreting Routing Statistics}
\label{app:routing}

\begin{table}[h]
\centering
\caption{
Average interpolation mass on the validation corpora (\%). These values measure
the contribution of each pathway to the injected residual and should not be
interpreted as discrete source-selection frequencies.
}
\label{tab:routing-stats}
\begin{tabular}{lrrr}
\toprule
Validation corpus & E & GE & GH\\
\midrule
Wikipedia-2021, QA checkpoint & 23.59 & 7.56 & 68.85\\
General mixture, NLP checkpoint & 62.85 & 4.49 & 32.67\\
Nemotron-CC-Code, coding checkpoint & 64.80 & 5.11 & 30.09\\
\bottomrule
\end{tabular}
\end{table}

Table~\ref{tab:routing-stats} reports the average effective contribution of
E, GE, and GH to the injected memory residual on the corresponding validation
corpora. The statistic is averaged over token positions, layers, and batches.
Because the deployed update has the form
\[
r=(1-\alpha)e+\alpha g,
\]
E retains a contribution of $1-\alpha$ even when a generated pathway is
admitted. Thus, interpolation mass and source-selection frequency measure
different aspects of the routing behavior.

The QA checkpoint places most of its interpolation mass on GH, whereas the NLP
and coding checkpoints retain substantially more mass on E. GE receives a
smaller average contribution in all three settings. This variation suggests
that the learned balance among the three pathways depends strongly on the
training domain and checkpoint.

For comparison, the calibration teacher stream records generated-pathway
admission rates of 99.97\% for QA and 98.07\% for NLP. These high admission
rates are compatible with non-trivial E mass because an admitted generated
pathway can still be interpolated with E using $\alpha<1$. Admission rate and
interpolation mass therefore should not be conflated.

A more detailed downstream analysis could additionally report, for each task,
the fraction of positions admitting GE or GH, the exact-E fallback rate, the
conditional mean interpolation coefficient $\alpha$, and the resulting
effective pathway weights.
\section{Transfer and secondary evaluations}
\label{app:transfer}

\paragraph{Transfer.}
Mistral-to-Llama QA scores are 29.98, 36.40, 67.39, 30.41, and 25.17. WebQA exceeds the source Mistral router, and the other four are lower. On six NLP tasks, the target router mean is 40.48 versus 38.23 for target E, with Yahoo at 10\%. These results demonstrate target-side adaptation only. A matched advantage over bare Llama would require its same-scorer baseline, which is missing (Appendix~\ref{app:transfer}).

\begin{table}[h]
\centering
\caption{Target-Llama NLP accuracy (\%). Without matched bare-Llama scores, these compare target memory endpoints only.}
\small
\begin{tabular}{lrrrrrrr}
\toprule
Condition & SST2 & MR & CR & RT & AGN & Yahoo & Mean\\
\midrule
Target E &50.92&44.85&50.00&50.00&23.62&10.00&38.23\\
Target router &49.08&47.95&58.35&52.44&25.05&10.00&40.48\\
\bottomrule
\end{tabular}
\end{table}

The router improves MR, CR, RT, and AGN, decreases SST2, and leaves Yahoo unchanged. Low absolute accuracies make these interface-transfer diagnostics rather than evidence of broadly successful NLP transfer. Without the bare target model, positive transfer and avoidance of negative transfer are not established.
\begin{table}[h]
\centering
\caption{
Performance on the HaluEval benchmark for question answering and summarization.
Results report accuracy (\%). Small signed values denote percentage-point
differences from Mistral-7B-v0.3. Average is the unweighted mean of QA and
Summarization. RAG is not evaluated on summarization because this task requires
only the source document.
}
\label{tab:halueval}
\small
\setlength{\tabcolsep}{6pt}

\begin{tabular}{lrrr}
\toprule
Method
& QA $\uparrow$
& Summarization $\uparrow$
& Average $\uparrow$\\
\midrule

\groupbar{4}{Reported baselines}

Mistral-7B-v0.3
& 53.99
& 50.27
& 52.13\\

CPT
& \scorechange{46.49}{-7.50}
& \scorechange{47.39}{-2.88}
& 46.94\\

LoRA
& \scorechange{50.02}{-3.97}
& \scorechange{50.38}{+0.11}
& 50.20\\

RAG
& \scorechange{65.09}{+11.10}
& --
& --\\

MLP Memory
& \scorechange{64.07}{+10.08}
& \scorechange{52.41}{+2.14}
& 58.24\\

\midrule
\groupbar{4}{This study}

E only
& \scorechange{50.12}{-3.87}
& \scorechange{54.82}{+4.55}
& 52.47\\

\rowcolor{tablefocus}
\textbf{\textsc{MemoryAthena}}
& \scorechange{\textbf{49.54}}{-4.45}
& \scorechange{\textbf{74.83}}{+24.56}
& \textbf{62.19}\\

\bottomrule
\end{tabular}
\end{table}
HaluEval evaluates hallucination recognition \citep{li2023halueval}. Summarization improves by 20.01 points, while dialogue changes modestly and QA decreases. Confusion matrices and class-balance checks are needed before attributing the summarization gain to improved detection rather than response bias. Classification on this benchmark is not a direct measurement of hallucinations in free-form generation.
\section{Scaling Analysis}
\label{app:scaling}

We study two complementary scaling dimensions:
\emph{model scaling}, where the backbone and the complete memory-side system
are enlarged jointly, and \emph{training-token scaling}, where the architecture
is fixed and only the optimization budget is increased.
We report only completed runs.

\paragraph{Model-scaling setup.}
We use GPT-2 Small, Medium, Large, and XL as the backbone family.
As the backbone grows, the Engram table, generated-memory modules, readers,
and routing head are scaled jointly rather than varying the memory table in
isolation. Table~\ref{tab:scaling-config} summarizes the corresponding
architectures and parameter counts.

\begin{table}[t]
\centering
\caption{
Model-scaling configurations. ``Gen.'' reports generator
width/depth/number of generated latents. Memory-side total includes the Engram
table, generated-memory modules, readers, and router, but excludes the backbone.
}
\label{tab:scaling-config}
\small
\setlength{\tabcolsep}{4.5pt}
\renewcommand{\arraystretch}{1.08}
\begin{tabular}{@{}lrrrrrr@{}}
\toprule
\textbf{Scale}
& \textbf{Backbone}
& \textbf{Engram}
& \textbf{Gen.}
& \textbf{Readers}
& \textbf{Router}
& \textbf{Memory-side} \\
\midrule
Small
& 124M
& 33.554M
& 256/2/4
& 16
& 64
& \textbf{37.573M} \\

Medium
& 345M
& 93.716M
& 428/2/6
& 24
& 104
& \textbf{104.008M} \\

Large
& 774M
& 209.715M
& 640/3/8
& 40
& 160
& \textbf{238.212M} \\

XL
& 1.5B
& 405.537M
& 896/4/12
& 56
& 224
& \textbf{472.912M} \\
\bottomrule
\end{tabular}
\end{table}

The Engram addressing structure is kept fixed across scales, with maximum
$n$-gram order 3 and four heads per order. Generator width increases from
256 to 896, generator depth from 2 to 4 layers, the number of generated
latents from 4 to 12, the source-adaptation rank from 16 to 56, and the router
hidden width from 64 to 224. The resulting memory-side system grows from
37.6M parameters with GPT-2 Small to 472.9M with GPT-2 XL.

The routing head itself remains comparatively small. Its parameter count grows
from approximately 0.054M, 0.119M, and 0.233M to 0.413M across the four scales,
while most memory-side capacity is allocated to the Engram table and the
generated-memory interface.

\paragraph{Training procedure.}
We consider two corpora for model scaling. The WikiText setting uses a maximum
budget of 100M processed tokens per optimization stage, while the general-text
mixture uses up to 600M tokens per stage.

Training follows the same staged procedure as the main experiments.
The memory is first learned with the source backbone fixed. The
generated-memory modules and readers are then optimized while the backbone and
memory table are fixed. Finally, these components are frozen and only the
routing head is trained from E-relative advantage supervision.
The no-memory results are obtained by directly evaluating the corresponding
pretrained GPT-2 backbones and are used only as reference points.

\begin{figure}[t]
    \centering

    \begin{subfigure}[t]{0.48\linewidth}
        \centering
        \includegraphics[width=\linewidth]
        {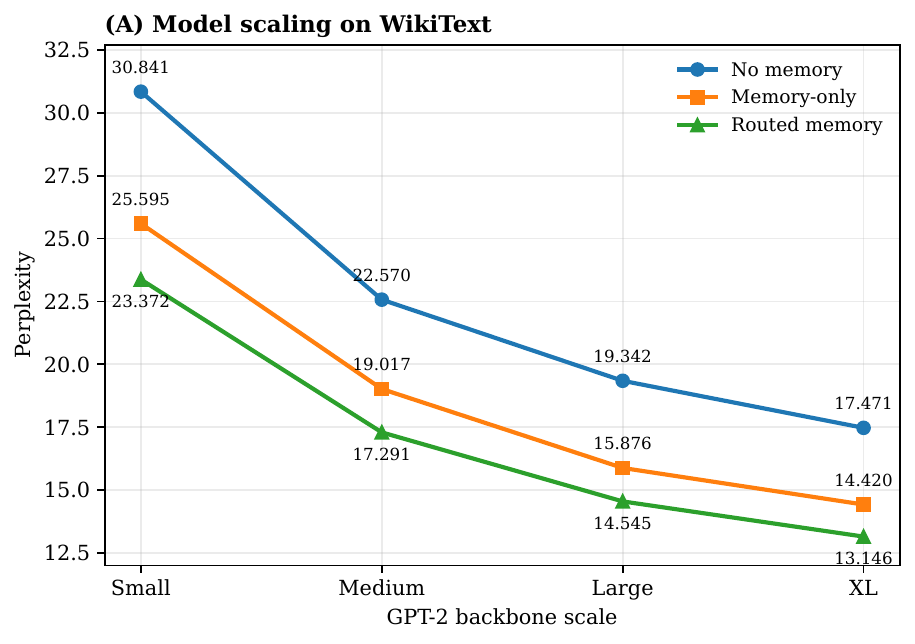}
        \caption{WikiText.}
        \label{fig:scaling-wikitext}
    \end{subfigure}
    \hfill
    \begin{subfigure}[t]{0.48\linewidth}
        \centering
        \includegraphics[width=\linewidth]
        {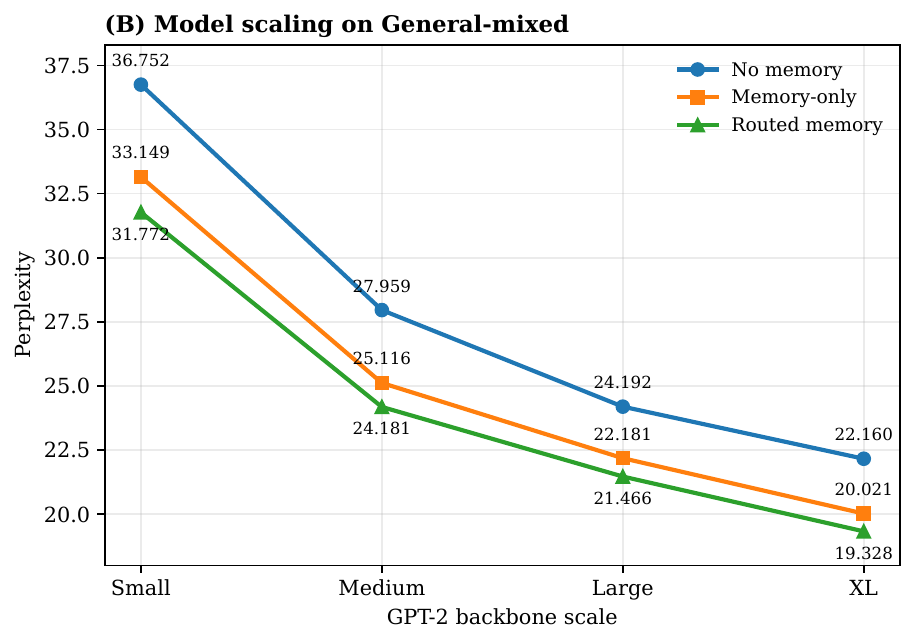}
        \caption{General-text mixture.}
        \label{fig:scaling-general}
    \end{subfigure}

    \caption{
    Model scaling across GPT-2 backbones. Lower perplexity is better.
    The routed system consistently improves over the corresponding memory-only
    configuration at every completed scale.
    }
    \label{fig:model-scaling}
\end{figure}

\paragraph{Model scaling.}
Figure~\ref{fig:model-scaling} shows the completed model-scaling experiments.
Across all completed runs, the same ordering is observed:
\[
\text{Router} < \text{Memory} < \text{No memory},
\]
where lower perplexity is better.

On WikiText, perplexity decreases from 30.841 to 24.033 to 23.372 for
GPT-2 Small, from 22.569 to 17.797 to 17.291 for Medium, and from
19.342 to 14.955 to 14.545 for Large, corresponding to the
no-memory, memory-only, and routed systems.

The completed general-mixture runs exhibit the same pattern. For GPT-2 Small,
perplexity decreases from 36.752 to 32.300 and then to 31.772; for Medium,
it decreases from 27.958 to 24.445 and then to 24.181.
Thus, the routing benefit persists as the backbone and memory-side system are
jointly scaled. The current results support persistence of the gain across
scale rather than an increasing routing advantage with model size.

\begin{table}[t]
\centering
\caption{
Exact perplexities for the completed model-scaling runs.
Lower is better.
}
\label{tab:model-scaling-values}
\small
\setlength{\tabcolsep}{7pt}
\begin{tabular}{@{}llrrr@{}}
\toprule
\textbf{Corpus}
& \textbf{Scale}
& \textbf{No memory}
& \textbf{Memory}
& \textbf{Router} \\
\midrule
WikiText
& Small
& 30.841
& 24.033
& \textbf{23.372} \\

& Medium
& 22.569
& 17.797
& \textbf{17.291} \\

& Large
& 19.342
& 14.955
& \textbf{14.545} \\

\addlinespace[2pt]
General mixture
& Small
& 36.752
& 32.300
& \textbf{31.772} \\

& Medium
& 27.958
& 24.445
& \textbf{24.181} \\
\bottomrule
\end{tabular}
\end{table}

\paragraph{Training-token scaling.}
We next isolate the effect of training budget while holding model capacity
fixed. These experiments use GPT-2 XL with the same 405.5M-parameter Engram
table and identical generator, reader, and router architectures. Only the
number of processed training tokens per stage is varied.

\begin{figure}[t]
    \centering

    \begin{subfigure}[t]{0.44\linewidth}
        \centering
        \includegraphics[width=\linewidth]
        {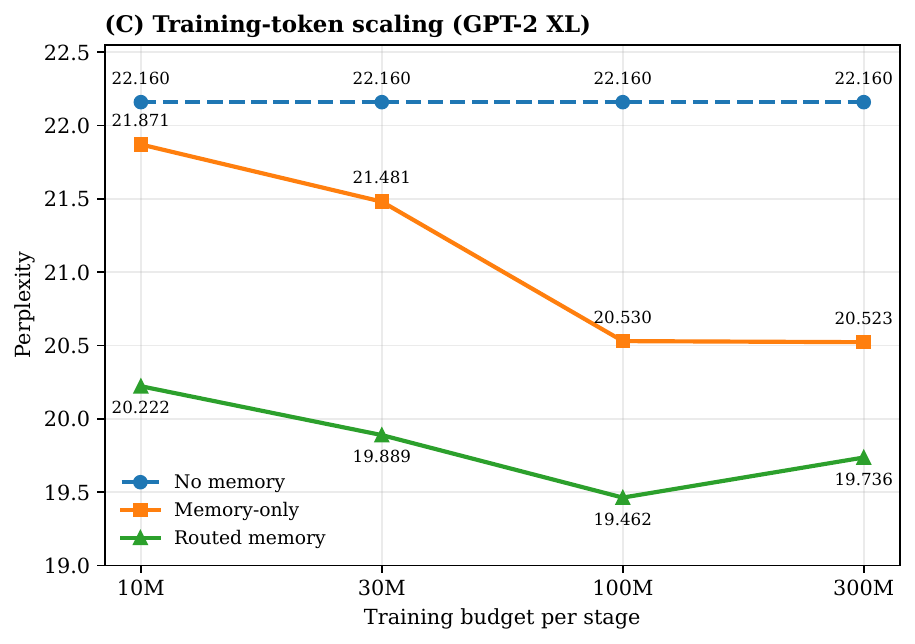}
        \caption{Training-token scaling with GPT-2 XL fixed.}
        \label{fig:scaling-token}
    \end{subfigure}
    \hfill
    \begin{subfigure}[t]{0.51\linewidth}
        \centering
        \includegraphics[width=\linewidth]
        {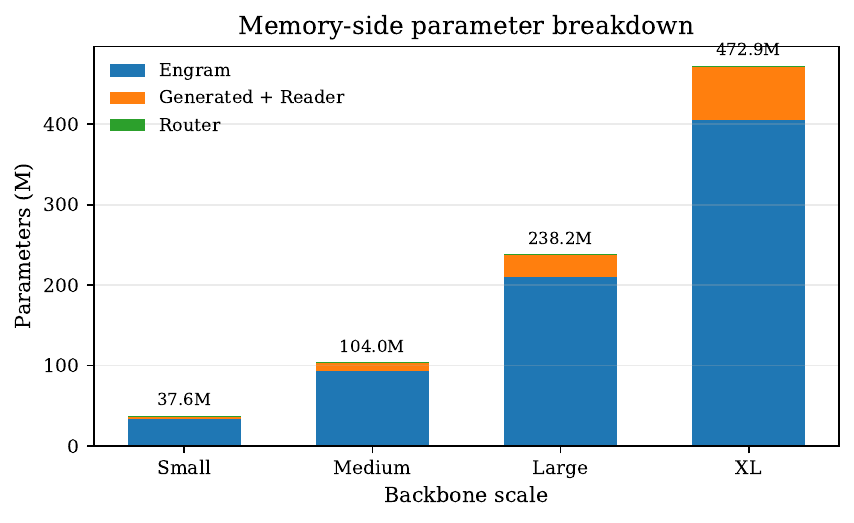}
        \caption{Memory-side parameter scaling.}
        \label{fig:scaling-params}
    \end{subfigure}

    \caption{
    Training and capacity scaling. Left: increasing the per-stage training
    budget improves both memory-only and routed systems while routing remains
    consistently better. Right: breakdown of memory-side parameters as the
    complete memory system is scaled with the backbone.
    }
    \label{fig:token-param-scaling}
\end{figure}

Figure~\ref{fig:token-param-scaling}a shows that increasing the training budget
from 10M to 30M and 100M tokens monotonically reduces perplexity for both
systems. Memory-only perplexity decreases from 20.714 to 20.369 and 19.783,
while routed-memory perplexity decreases from 20.222 to 19.889 and 19.462.
The routed model therefore remains better than memory alone at every completed
training budget.

Figure~\ref{fig:token-param-scaling}b provides the corresponding parameter
breakdown across model scales. Most of the memory-side capacity is allocated
to the Engram table, followed by the generated-memory modules and readers,
while the routing head contributes only a small fraction of the total
parameter count.

\begin{table}[h]
\centering
\caption{
Training-token scaling with the GPT-2 XL architecture fixed.
Only the per-stage optimization budget changes.
}
\label{tab:token-scaling}
\small
\setlength{\tabcolsep}{10pt}
\begin{tabular}{@{}rrr@{}}
\toprule
\textbf{Training budget}
& \textbf{Memory}
& \textbf{Router} \\
\midrule
10M
& 20.714
& \textbf{20.222} \\

30M
& 20.369
& \textbf{19.889} \\

100M
& 19.783
& \textbf{19.462} \\
\bottomrule
\end{tabular}
\end{table}

Overall, the completed scaling experiments show two consistent trends.
First, the benefit of adaptive routing is preserved as the backbone and
memory-side architecture are jointly enlarged. Second, increasing the training
budget improves both memory-only and routed systems, while the routing gain
remains present throughout the evaluated range.

\section{Case Study}
\label{sec:case-study}

\begin{figure*}[t]
    \centering
    \includegraphics[width=0.96\textwidth]{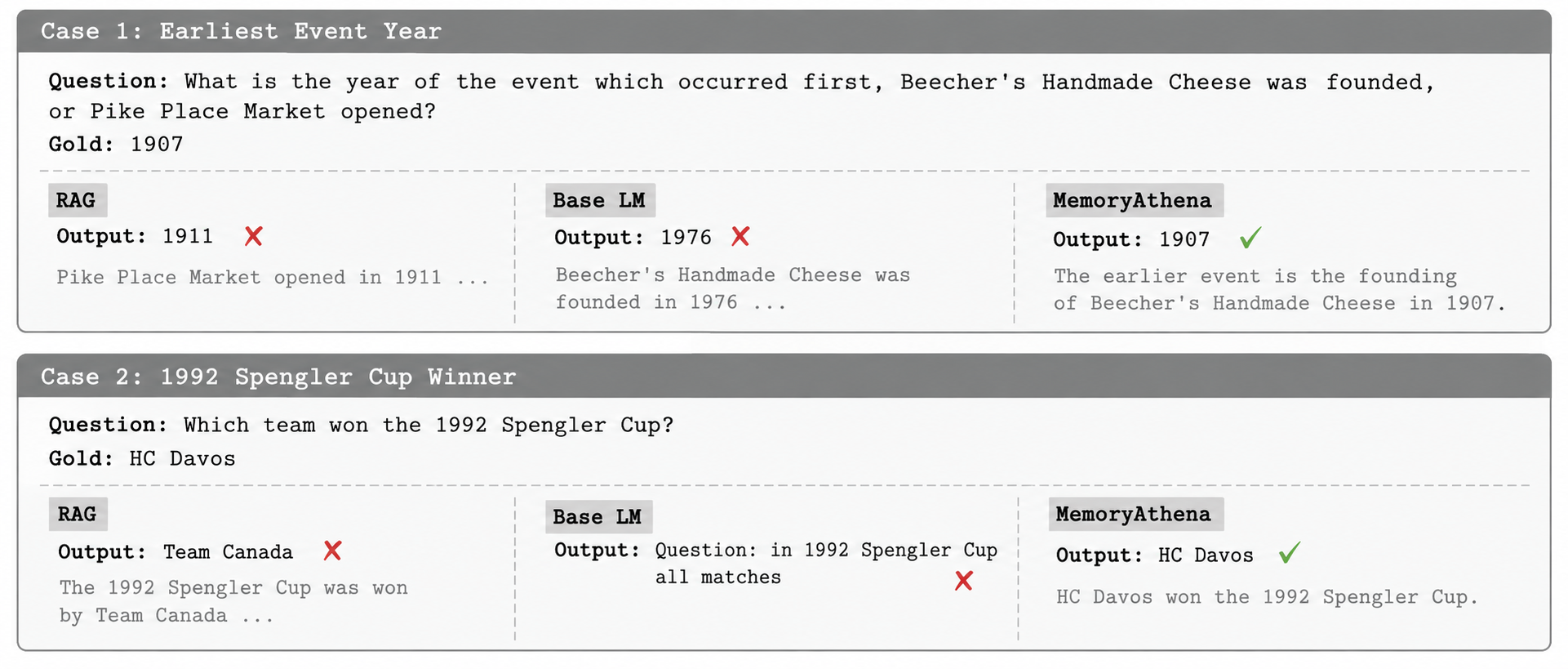}
    \caption{
    Two downstream case studies of \textsc{MemoryAthena}.
    The routed system is correct in both examples although all individual
    endpoints ($E$, $GE$, and $GH$) are incorrect.
    The route traces show the effective source mass assigned to each pathway.
    }
    \label{fig:case-study}
\end{figure*}

Figure~\ref{fig:case-study} shows two representative downstream examples.
In both cases, none of the standalone pathways ($E$, $GE$, or $GH$) yields the
correct final answer, whereas \textsc{MemoryAthena} does.

In the first example, the task is to identify which of two events occurred
earlier.
Although $GH$ receives most of the routed source mass, its standalone answer is
still incorrect, while the routed system recovers the correct year, 1907.
In the second example, the question asks for the winner of the 1992 Spengler
Cup.
The standalone outputs are either noisy or incomplete, but the routed system
returns the exact answer, \textit{HC Davos}.
These examples illustrate that the benefit of routing is not simply selecting
the best single endpoint.
Instead, \textsc{MemoryAthena} can exploit complementary information across
memory pathways and transform imperfect endpoint predictions into a correct
final answer.

\section{Computational Cost}
\label{app:compute}

We analyze the computational cost of \textsc{MemoryAthena} from two
perspectives: the additional optimization required to learn the generated
memory interface and routing policy, and the runtime overhead introduced during
inference. The routing head itself is parameter-light, but the complete system
must additionally evaluate generated-memory components and, in the current
implementation, obtain clean causal backbone states for the GH pathway.

\subsection{Training Cost}

Training follows the three-stage procedure described in
Appendix~\ref{app:training}. The memory is learned first, the generated-memory
modules and readers are then optimized with the backbone and memory table
frozen, and the final stage trains only the E-relative routing heads.

The reader/generator stage contains approximately 167.4M trainable parameters,
whereas the final router contains only 534,924 trainable parameters.
Thus, the routing head accounts for only a small fraction of the trainable
memory-interface parameters.

\begin{table}[t]
\centering
\caption{
Recorded wall-clock training time for the reader/generator and routing stages.
The router contains 0.535M trainable parameters, compared with approximately
167.4M in the reader/generator stage.
}
\label{tab:training-cost}
\small
\setlength{\tabcolsep}{6pt}
\begin{tabular}{@{}llrr@{}}
\toprule
\textbf{Setting}
& \textbf{Stage}
& \textbf{Trainable params.}
& \textbf{Time (h)} \\
\midrule
QA
& Reader / generator
& 167.386M
& 31.70 \\
& Router
& 0.535M
& 11.19 \\
\addlinespace

General NLP
& Reader / generator
& 167.365M
& 16.01 \\
& Router
& 0.535M
& 10.58 \\
\addlinespace

Coding
& Reader / generator
& 167.365M
& 16.41 \\
& Router
& 0.535M
& 10.94 \\
\bottomrule
\end{tabular}
\end{table}

The recorded reader/generator and router stages together take approximately
42.89 hours for QA, 26.59 hours for general NLP, and 27.35 hours for coding.
These numbers describe the recorded stages rather than the complete lifetime
cost of the reusable source memory.

Although only a small routing head is optimized in the final stage, routing
training still requires non-trivial computation. Counterfactual supervision is
constructed by evaluating multiple frozen endpoints under teacher forcing, and
the GH pathway additionally requires clean backbone states. Consequently,
trainable parameter count alone is not a direct measure of total training
compute.

\subsection{Inference Microbenchmark}

We additionally benchmark the inference overhead of \textsc{MemoryAthena}
relative to the direct E-only pathway across GPT-2 Small, Medium, Large, and XL.

All measurements are performed on an AMD MI250X GPU using BF16 precision.
Each run uses a fixed synthetic sequence of 338 input tokens and generates
16 output tokens. We perform one warmup iteration followed by three measured
iterations. We report end-to-end latency, total-token throughput, and peak
reserved GPU memory.

The E-only configuration executes the direct Engram pathway. In contrast,
\textsc{MemoryAthena} additionally evaluates the generated-memory interface,
routing features, and the clean causal backbone states required by GH in the
current implementation.

\begin{figure*}[t]
    \centering

    \begin{subfigure}[t]{0.32\textwidth}
        \centering
        \includegraphics[width=\linewidth]
        {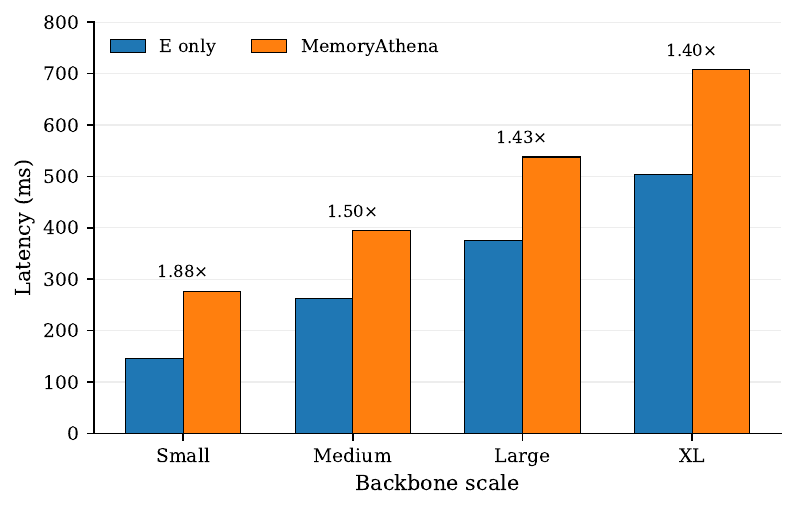}
        \caption{End-to-end latency.}
        \label{fig:compute-latency}
    \end{subfigure}
    \hfill
    \begin{subfigure}[t]{0.32\textwidth}
        \centering
        \includegraphics[width=\linewidth]
        {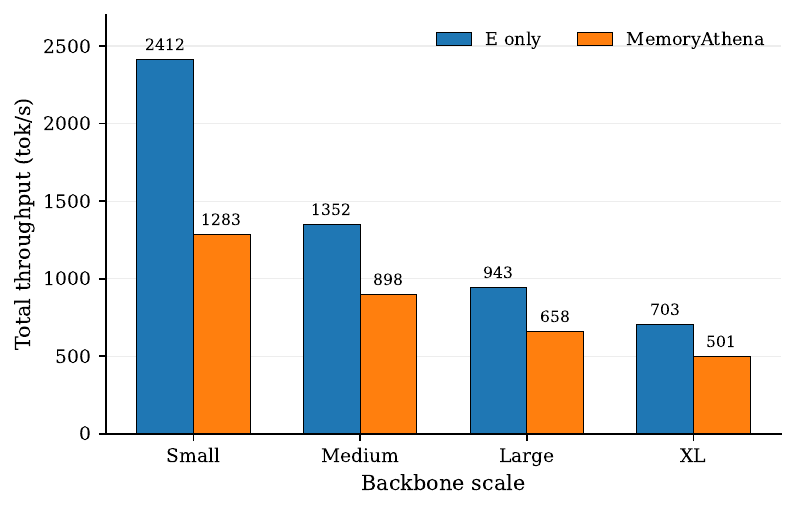}
        \caption{Total-token throughput.}
        \label{fig:compute-throughput}
    \end{subfigure}
    \hfill
    \begin{subfigure}[t]{0.32\textwidth}
        \centering
        \includegraphics[width=\linewidth]
        {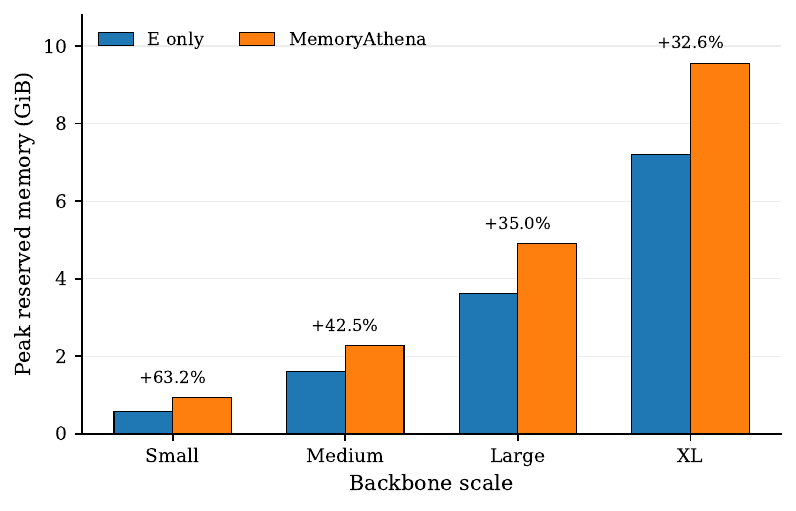}
        \caption{Peak reserved GPU memory.}
        \label{fig:compute-memory}
    \end{subfigure}

    \caption{
    Inference microbenchmark on an AMD MI250X using BF16, with a fixed
    338-token input and 16 generated tokens.
    \textbf{Left:} latency, where annotations show the speed advantage of
    E-only inference.
    \textbf{Middle:} total-token throughput.
    \textbf{Right:} peak reserved GPU memory, with annotations showing the
    relative memory overhead of \textsc{MemoryAthena}.
    The relative runtime and memory overhead decrease as the backbone scales.
    }
    \label{fig:compute-microbenchmark}
\end{figure*}

\paragraph{Latency and throughput.}
E-only inference is faster at all four evaluated scales.
For GPT-2 Small, latency increases from 146.7\,ms for E-only to 276.0\,ms for
\textsc{MemoryAthena}, corresponding to a 1.88$\times$ speed advantage for
E-only. The relative gap decreases with model size, to 1.50$\times$ for
Medium, 1.43$\times$ for Large, and 1.40$\times$ for XL.

The corresponding total-token throughput decreases from 2412.4 to
1282.9 tokens/s at Small, from 1352.0 to 898.4 tokens/s at Medium, from
942.7 to 658.4 tokens/s at Large, and from 702.6 to 500.7 tokens/s at XL.
Thus, the absolute cost of both systems increases with model scale, while the
relative overhead of the routed system becomes smaller.

\paragraph{Memory overhead.}
Peak reserved GPU memory increases from 0.57 to 0.93\,GiB at Small,
1.60 to 2.28\,GiB at Medium, 3.63 to 4.90\,GiB at Large, and
7.21 to 9.56\,GiB at XL.
These correspond to relative overheads of approximately 63.2\%, 42.5\%,
35.0\%, and 32.6\%, respectively.

The memory results therefore exhibit the same qualitative trend as latency:
although \textsc{MemoryAthena} requires additional runtime state, this
additional cost represents a smaller fraction of the overall system footprint
as the backbone becomes larger.

\paragraph{Where does the overhead come from?}
The additional cost should not be attributed primarily to the routing MLP.
The routing head contains only 534,924 parameters. Instead, the main runtime
overhead comes from evaluating the generated-memory pathways and maintaining
their intermediate states. In particular, GH currently requires a separate
memory-disabled backbone computation to obtain its clean causal conditioning
states.

This distinction is important: \textsc{MemoryAthena} is
\emph{parameter-efficient} as a routing mechanism, but it is not a
zero-overhead inference method.

\paragraph{Scope of the benchmark.}
The experiment is a controlled microbenchmark using a fixed synthetic prompt,
generation length, precision, and hardware configuration. Its purpose is to
measure relative systems overhead under matched conditions. The reported
numbers should therefore not be interpreted as full downstream-task
throughput, which can vary with sequence length, batch size, generation length,
routing behavior, and hardware utilization.

Overall, \textsc{MemoryAthena} trades additional computation for adaptive use
of generated memory. The inference overhead is measurable at all evaluated
scales, but its relative cost decreases with backbone size: the E-only latency
advantage falls from 1.88$\times$ at Small to 1.40$\times$ at XL, while the
reserved-memory overhead decreases from approximately 63\% to 33\%.
\section{Language-model diagnostics and incomplete extensions}
\label{app:secondary}

\begin{table}[h]
\centering
\caption{General-NLP within-run validation perplexity. Endpoints and downstream accuracy need not rank identically.}
\begin{tabular}{rrrr}
\toprule
E & GE & GH & Router\\
\midrule
9.07260&8.79091&8.78358&8.83063\\
\bottomrule
\end{tabular}
\end{table}

The NLP router improves E perplexity but not either generated endpoint. QA likewise records lower validation perplexity for ordinary soft fusion (7.07310) and hard routing (7.09269) than for the advantage router (7.14538), with E at 8.57136. Corpus perplexity alone does not establish better QA performance.

\paragraph{Coding.}
Completed Nemotron-CC-Code expert records give perplexities 3.10842 (E), 2.99503 (GE), and 2.98369 (GH), with arithmetic mean 3.02853. Routing records give 2.99563 after 19,998,720 input positions which is lower than tri-experts mean. 

%% file: arxiv.bbl
\begin{thebibliography}{26}
\providecommand{\natexlab}[1]{#1}
\providecommand{\url}[1]{\texttt{#1}}
\expandafter\ifx\csname urlstyle\endcsname\relax
  \providecommand{\doi}[1]{doi: #1}\else
  \providecommand{\doi}{doi: \begingroup \urlstyle{rm}\Url}\fi

\bibitem[Berant et~al.(2013)Berant, Chou, Frostig, and Liang]{berant2013semantic}
J.~Berant, A.~Chou, R.~Frostig, and P.~Liang.
\newblock Semantic parsing on {F}reebase from question-answer pairs.
\newblock In D.~Yarowsky, T.~Baldwin, A.~Korhonen, K.~Livescu, and S.~Bethard, editors, \emph{Proceedings of the 2013 Conference on Empirical Methods in Natural Language Processing}, pages 1533--1544, Seattle, Washington, USA, Oct. 2013. Association for Computational Linguistics.
\newblock URL \url{https://aclanthology.org/D13-1160/}.

\bibitem[Cheng et~al.(2026)Cheng, Zeng, Dai, Chen, Wang, Xie, Huang, Yu, Hao, Zhang, Li, Zhang, Zhao, and Liang]{cheng2026engram}
X.~Cheng, W.~Zeng, D.~Dai, Q.~Chen, B.~Wang, Z.~Xie, K.~Huang, X.~Yu, Z.~Hao, H.~Zhang, Y.-K. Li, H.~Zhang, D.~Zhao, and W.~Liang.
\newblock Conditional memory via scalable lookup: A new axis of sparsity for large language models.
\newblock In M.~Liakata, V.~P. Moreira, J.~Zhang, and D.~Jurgens, editors, \emph{Proceedings of the 64th Annual Meeting of the {A}ssociation for {C}omputational {L}inguistics (Volume 1: Long Papers)}, pages 4968--4990, San Diego, California, United States, July 2026. Association for Computational Linguistics.
\newblock \doi{10.18653/v1/2026.acl-long.226}.
\newblock URL \url{https://aclanthology.org/2026.acl-long.226/}.

\bibitem[Fedus et~al.(2022)Fedus, Zoph, and Shazeer]{fedus2022switch}
W.~Fedus, B.~Zoph, and N.~Shazeer.
\newblock Switch transformers: Scaling to trillion parameter models with simple and efficient sparsity.
\newblock \emph{Journal of Machine Learning Research}, 23\penalty0 (120):\penalty0 1--39, 2022.
\newblock URL \url{http://jmlr.org/papers/v23/21-0998.html}.

\bibitem[Feng et~al.(2025)Feng, Wang, Chen, and Han]{feng2025preresqr1}
Z.~Feng, W.~Wang, X.~Chen, and T.~Han.
\newblock {PreResQ-R1}: Response-preference disentangled ranking-and-scoring reinforcement optimization for robust visual quality assessment, 2025.
\newblock URL \url{https://arxiv.org/abs/2511.05393}.

\bibitem[Feng et~al.(2026)Feng, Zhu, Zhao, Zhao, Ren, and Han]{feng2026trimrg}
Z.~Feng, D.~Zhu, J.~Zhao, P.~Zhao, J.~Ren, and T.~Han.
\newblock {TriMRG}: Toward hallucination-resistant medical image report generation via triple graph semantic reasoning and rethinking.
\newblock \emph{IEEE Transactions on Circuits and Systems for Video Technology}, 2026.
\newblock \doi{10.1109/TCSVT.2026.3712885}.
\newblock URL \url{https://doi.org/10.1109/TCSVT.2026.3712885}.

\bibitem[Hamborg et~al.(2017)Hamborg, Meuschke, Breitinger, and Gipp]{blagojevic2021ccnews}
F.~Hamborg, N.~Meuschke, C.~Breitinger, and B.~Gipp.
\newblock news-please: A generic news crawler and extractor.
\newblock In \emph{Proceedings of the 15th International Symposium of Information Science}, pages 218--223, March 2017.
\newblock \doi{10.5281/zenodo.4120316}.

\bibitem[Holtzman et~al.(2021)Holtzman, West, Shwartz, Choi, and Zettlemoyer]{holtzman2021surface}
A.~Holtzman, P.~West, V.~Shwartz, Y.~Choi, and L.~Zettlemoyer.
\newblock Surface form competition: Why the highest probability answer isn{'}t always right.
\newblock In M.-F. Moens, X.~Huang, L.~Specia, and S.~W.-t. Yih, editors, \emph{Proceedings of the 2021 Conference on Empirical Methods in Natural Language Processing}, pages 7038--7051, Online and Punta Cana, Dominican Republic, Nov. 2021. Association for Computational Linguistics.
\newblock \doi{10.18653/v1/2021.emnlp-main.564}.
\newblock URL \url{https://aclanthology.org/2021.emnlp-main.564/}.

\bibitem[Hu and Liu(2004)]{hu2004mining}
M.~Hu and B.~Liu.
\newblock Mining and summarizing customer reviews.
\newblock In \emph{Proceedings of the tenth ACM SIGKDD international conference on Knowledge discovery and data mining}, KDD04, pages 168--177. ACM, Aug. 2004.
\newblock \doi{10.1145/1014052.1014073}.
\newblock URL \url{https://doi.org/10.1145/1014052.1014073}.

\bibitem[Jiang et~al.(2023)Jiang, Sablayrolles, Mensch, Bamford, Chaplot, de~las Casas, Bressand, Lengyel, Lample, Saulnier, Lavaud, Lachaux, Stock, Scao, Lavril, Wang, Lacroix, and Sayed]{jiang2023mistral}
A.~Q. Jiang, A.~Sablayrolles, A.~Mensch, C.~Bamford, D.~S. Chaplot, D.~de~las Casas, F.~Bressand, G.~Lengyel, G.~Lample, L.~Saulnier, L.~R. Lavaud, M.-A. Lachaux, P.~Stock, T.~L. Scao, T.~Lavril, T.~Wang, T.~Lacroix, and W.~E. Sayed.
\newblock Mistral {7B}, 2023.
\newblock URL \url{https://arxiv.org/abs/2310.06825}.

\bibitem[Joshi et~al.(2017)Joshi, Choi, Weld, and Zettlemoyer]{joshi2017trivia}
M.~Joshi, E.~Choi, D.~Weld, and L.~Zettlemoyer.
\newblock {T}rivia{QA}: A large scale distantly supervised challenge dataset for reading comprehension.
\newblock In R.~Barzilay and M.-Y. Kan, editors, \emph{Proceedings of the 55th Annual Meeting of the Association for Computational Linguistics (Volume 1: Long Papers)}, pages 1601--1611, Vancouver, Canada, July 2017. Association for Computational Linguistics.
\newblock \doi{10.18653/v1/P17-1147}.
\newblock URL \url{https://aclanthology.org/P17-1147/}.

\bibitem[Khandelwal et~al.(2020)Khandelwal, Levy, Jurafsky, Zettlemoyer, and Lewis]{khandelwal2020knn}
U.~Khandelwal, O.~Levy, D.~Jurafsky, L.~Zettlemoyer, and M.~Lewis.
\newblock Generalization through memorization: Nearest neighbor language models.
\newblock In \emph{International Conference on Learning Representations}, 2020.
\newblock URL \url{https://openreview.net/forum?id=HklBjCEKvH}.

\bibitem[Kwiatkowski et~al.(2019)Kwiatkowski, Palomaki, Redfield, Collins, Parikh, Alberti, Epstein, Polosukhin, Devlin, Lee, Toutanova, Jones, Kelcey, Chang, Dai, Uszkoreit, Le, and Petrov]{kwiatkowski2019natural}
T.~Kwiatkowski, J.~Palomaki, O.~Redfield, M.~Collins, A.~Parikh, C.~Alberti, D.~Epstein, I.~Polosukhin, J.~Devlin, K.~Lee, K.~Toutanova, L.~Jones, M.~Kelcey, M.-W. Chang, A.~M. Dai, J.~Uszkoreit, Q.~Le, and S.~Petrov.
\newblock Natural questions: A benchmark for question answering research.
\newblock \emph{Transactions of the Association for Computational Linguistics}, 7:\penalty0 452--466, 2019.
\newblock \doi{10.1162/tacl_a_00276}.
\newblock URL \url{https://aclanthology.org/Q19-1026/}.

\bibitem[Lewis et~al.(2020)Lewis, Perez, Piktus, Petroni, Karpukhin, Goyal, K\"{u}ttler, Lewis, Yih, Rockt\"{a}schel, Riedel, and Kiela]{lewis2020rag}
P.~Lewis, E.~Perez, A.~Piktus, F.~Petroni, V.~Karpukhin, N.~Goyal, H.~K\"{u}ttler, M.~Lewis, W.-t. Yih, T.~Rockt\"{a}schel, S.~Riedel, and D.~Kiela.
\newblock Retrieval-augmented generation for knowledge-intensive {NLP} tasks.
\newblock In H.~Larochelle, M.~Ranzato, R.~Hadsell, M.~Balcan, and H.~Lin, editors, \emph{Advances in Neural Information Processing Systems}, volume~33, pages 9459--9474. Curran Associates, Inc., 2020.
\newblock URL \url{https://proceedings.neurips.cc/paper_files/paper/2020/file/6b493230205f780e1bc26945df7481e5-Paper.pdf}.

\bibitem[Li et~al.(2023)Li, Cheng, Zhao, Nie, and Wen]{li2023halueval}
J.~Li, X.~Cheng, X.~Zhao, J.-Y. Nie, and J.-R. Wen.
\newblock {H}alu{E}val: A large-scale hallucination evaluation benchmark for large language models.
\newblock In H.~Bouamor, J.~Pino, and K.~Bali, editors, \emph{Proceedings of the 2023 Conference on Empirical Methods in Natural Language Processing}, pages 6449--6464, Singapore, Dec. 2023. Association for Computational Linguistics.
\newblock \doi{10.18653/v1/2023.emnlp-main.397}.
\newblock URL \url{https://aclanthology.org/2023.emnlp-main.397/}.

\bibitem[Li et~al.(2026)Li, Yu, Wang, and Ji]{li2026transfer}
M.~Li, G.~Yu, X.~Wang, and S.~Ji.
\newblock Cross-model memory transfer via target-side reader adaptation, 2026.
\newblock URL \url{https://arxiv.org/abs/2608.17050}.

\bibitem[Lin et~al.(2022)Lin, Hilton, and Evans]{lin2022truthful}
S.~Lin, J.~Hilton, and O.~Evans.
\newblock {T}ruthful{QA}: Measuring how models mimic human falsehoods.
\newblock In S.~Muresan, P.~Nakov, and A.~Villavicencio, editors, \emph{Proceedings of the 60th Annual Meeting of the Association for Computational Linguistics (Volume 1: Long Papers)}, pages 3214--3252, Dublin, Ireland, May 2022. Association for Computational Linguistics.
\newblock \doi{10.18653/v1/2022.acl-long.229}.
\newblock URL \url{https://aclanthology.org/2022.acl-long.229/}.

\bibitem[Maas et~al.(2011)Maas, Daly, Pham, Huang, Ng, and Potts]{maas2011learning}
A.~L. Maas, R.~E. Daly, P.~T. Pham, D.~Huang, A.~Y. Ng, and C.~Potts.
\newblock Learning word vectors for sentiment analysis.
\newblock In D.~Lin, Y.~Matsumoto, and R.~Mihalcea, editors, \emph{Proceedings of the 49th Annual Meeting of the Association for Computational Linguistics: Human Language Technologies}, pages 142--150, Portland, Oregon, USA, June 2011. Association for Computational Linguistics.
\newblock URL \url{https://aclanthology.org/P11-1015/}.

\bibitem[Merity et~al.(2017)Merity, Xiong, Bradbury, and Socher]{merity2017pointer}
S.~Merity, C.~Xiong, J.~Bradbury, and R.~Socher.
\newblock Pointer sentinel mixture models.
\newblock In \emph{International Conference on Learning Representations}, 2017.
\newblock URL \url{https://openreview.net/forum?id=Byj72udxe}.

\bibitem[Pang and Lee(2005)]{pang2005seeing}
B.~Pang and L.~Lee.
\newblock Seeing stars: Exploiting class relationships for sentiment categorization with respect to rating scales.
\newblock In K.~Knight, H.~T. Ng, and K.~Oflazer, editors, \emph{Proceedings of the 43rd Annual Meeting of the Association for Computational Linguistics ({ACL}{'}05)}, pages 115--124, Ann Arbor, Michigan, June 2005. Association for Computational Linguistics.
\newblock \doi{10.3115/1219840.1219855}.
\newblock URL \url{https://aclanthology.org/P05-1015/}.

\bibitem[Shazeer et~al.(2017)Shazeer, Mirhoseini, Maziarz, Davis, Le, Hinton, and Dean]{shazeer2017moe}
N.~Shazeer, A.~Mirhoseini, K.~Maziarz, A.~Davis, Q.~Le, G.~Hinton, and J.~Dean.
\newblock Outrageously large neural networks: The sparsely-gated mixture-of-experts layer.
\newblock In \emph{International Conference on Learning Representations}, 2017.
\newblock URL \url{https://openreview.net/forum?id=B1ckMDqlg}.

\bibitem[Socher et~al.(2013)Socher, Perelygin, Wu, Chuang, Manning, Ng, and Potts]{socher2013recursive}
R.~Socher, A.~Perelygin, J.~Wu, J.~Chuang, C.~D. Manning, A.~Ng, and C.~Potts.
\newblock Recursive deep models for semantic compositionality over a sentiment treebank.
\newblock In D.~Yarowsky, T.~Baldwin, A.~Korhonen, K.~Livescu, and S.~Bethard, editors, \emph{Proceedings of the 2013 Conference on Empirical Methods in Natural Language Processing}, pages 1631--1642, Seattle, Washington, USA, Oct. 2013. Association for Computational Linguistics.
\newblock URL \url{https://aclanthology.org/D13-1170/}.

\bibitem[Touvron et~al.(2023)Touvron, Martin, Stone, Albert, Almahairi, Babaei, Bashlykov, Batra, Bhargava, Bhosale, Bikel, Blecher, Ferrer, Chen, Cucurull, Esiobu, Fernandes, Fu, Fu, Fuller, Gao, Goswami, Goyal, Hartshorn, Hosseini, Hou, Inan, Kardas, Kerkez, Khabsa, Kloumann, Korenev, Koura, Lachaux, Lavril, Lee, Liskovich, Lu, Mao, Martinet, Mihaylov, Mishra, Molybog, Nie, Poulton, Reizenstein, Rungta, Saladi, Schelten, Silva, Smith, Subramanian, Tan, Tang, Taylor, Williams, Kuan, Xu, Yan, Zarov, Zhang, Fan, Kambadur, Narang, Rodriguez, Stojnic, Edunov, and Scialom]{touvron2023llama}
H.~Touvron, L.~Martin, K.~Stone, P.~Albert, A.~Almahairi, Y.~Babaei, N.~Bashlykov, S.~Batra, P.~Bhargava, S.~Bhosale, D.~Bikel, L.~Blecher, C.~C. Ferrer, M.~Chen, G.~Cucurull, D.~Esiobu, J.~Fernandes, J.~Fu, W.~Fu, B.~Fuller, C.~Gao, V.~Goswami, N.~Goyal, A.~Hartshorn, S.~Hosseini, R.~Hou, H.~Inan, M.~Kardas, V.~Kerkez, M.~Khabsa, I.~Kloumann, A.~Korenev, P.~S. Koura, M.-A. Lachaux, T.~Lavril, J.~Lee, D.~Liskovich, Y.~Lu, Y.~Mao, X.~Martinet, T.~Mihaylov, P.~Mishra, I.~Molybog, Y.~Nie, A.~Poulton, J.~Reizenstein, R.~Rungta, K.~Saladi, A.~Schelten, R.~Silva, E.~M. Smith, R.~Subramanian, X.~E. Tan, B.~Tang, R.~Taylor, A.~Williams, J.~X. Kuan, P.~Xu, Z.~Yan, I.~Zarov, Y.~Zhang, A.~Fan, M.~Kambadur, S.~Narang, A.~Rodriguez, R.~Stojnic, S.~Edunov, and T.~Scialom.
\newblock Llama 2: Open foundation and fine-tuned chat models, 2023.
\newblock URL \url{https://arxiv.org/abs/2307.09288}.

\bibitem[Wei et~al.(2026)Wei, Cao, Wang, Kai, Guo, Zhou, and Lin]{wei2025mlp}
R.~Wei, J.~Cao, J.~Wang, J.~Kai, Q.~Guo, B.~Zhou, and Z.~Lin.
\newblock {MLP} memory: A retriever-pretrained memory for large language models.
\newblock In \emph{International Conference on Learning Representations}, 2026.
\newblock URL \url{https://openreview.net/forum?id=1SMdxRtLBp}.

\bibitem[Xie et~al.(2026)Xie, Zhao, Liu, Zhu, Chen, Ye, Chen, Xu, Dong, Wang, et~al.]{xie2026step}
W.~Xie, H.~Zhao, W.~Liu, Y.~Zhu, L.~Chen, M.~Ye, Z.~Chen, Y.~Xu, S.~Dong, Z.~Wang, et~al.
\newblock Step-wise rubric rewards for llm reasoning.
\newblock \emph{arXiv preprint arXiv:2605.17291}, 2026.

\bibitem[Yang et~al.(2018)Yang, Qi, Zhang, Bengio, Cohen, Salakhutdinov, and Manning]{yang2018hotpot}
Z.~Yang, P.~Qi, S.~Zhang, Y.~Bengio, W.~Cohen, R.~Salakhutdinov, and C.~D. Manning.
\newblock {H}otpot{QA}: A dataset for diverse, explainable multi-hop question answering.
\newblock In E.~Riloff, D.~Chiang, J.~Hockenmaier, and J.~Tsujii, editors, \emph{Proceedings of the 2018 Conference on Empirical Methods in Natural Language Processing}, pages 2369--2380, Brussels, Belgium, Oct.-Nov. 2018. Association for Computational Linguistics.
\newblock \doi{10.18653/v1/D18-1259}.
\newblock URL \url{https://aclanthology.org/D18-1259/}.

\bibitem[Zhang et~al.(2015)Zhang, Zhao, and LeCun]{zhang2015character}
X.~Zhang, J.~Zhao, and Y.~LeCun.
\newblock Character-level convolutional networks for text classification.
\newblock In C.~Cortes, N.~Lawrence, D.~Lee, M.~Sugiyama, and R.~Garnett, editors, \emph{Advances in Neural Information Processing Systems}, volume~28. Curran Associates, Inc., 2015.
\newblock URL \url{https://proceedings.neurips.cc/paper_files/paper/2015/file/250cf8b51c773f3f8dc8b4be867a9a02-Paper.pdf}.

\end{thebibliography}
